\documentclass{article}

\usepackage{graphicx} 
\usepackage[numbers]{natbib} 

\usepackage{amsmath}
\usepackage{amsfonts}
\usepackage{adjustbox}
\usepackage{comment}
\usepackage[margin=1in]{geometry}
\usepackage{bm}
\usepackage{url}
\usepackage{authblk}

\usepackage{booktabs}
\usepackage{longtable}
\usepackage{array}
\usepackage{multirow}
\usepackage{wrapfig}
\usepackage{float}
\usepackage{colortbl}
\usepackage{pdflscape}
\usepackage{tabu}
\usepackage{threeparttable}
\usepackage{threeparttablex}
\usepackage[normalem]{ulem}
\usepackage{makecell}
\usepackage{xcolor}
\usepackage{booktabs}
\usepackage{longtable}
\usepackage{array}
\usepackage{multirow}
\usepackage{wrapfig}
\usepackage{float}
\usepackage{colortbl}
\usepackage{pdflscape}
\usepackage{tabu}
\usepackage{threeparttable}
\usepackage{threeparttablex}
\usepackage[normalem]{ulem}
\usepackage{makecell}
\usepackage{xcolor}

\usepackage[T1]{fontenc}
\usepackage{lmodern}        
\usepackage[utf8]{inputenc} 

\title{The Sensitivity of Variational Bayesian Neural Network Performance to Hyperparameters}

\author[1]{Scott Koermer \thanks{Corresponding author, skoermer@lanl.gov}}
\author[1]{Natalie Klein}

\affil[1]{Statistical Sciences Group, Los Alamos National Laboratory, Los Alamos, NM, USA}

\date{\today}

\begin{document}

\maketitle

\begin{abstract}
In scientific applications, predictive modeling is often of limited use without accurate uncertainty quantification (UQ) to indicate when a model may be extrapolating or when more data needs to be collected.
Bayesian Neural Networks (BNNs) produce predictive uncertainty by propagating uncertainty in neural network (NN) weights and offer the promise of obtaining not only an accurate predictive model but also accurate UQ.
However, in practice, obtaining accurate UQ with BNNs is difficult due in part to the approximations used for practical model training and in part to the need to choose a suitable set of hyperparameters; these hyperparameters outnumber those needed for traditional NNs and often have opaque effects on the results.
We aim to shed light on the effects of hyperparameter choices for BNNs by performing a global sensitivity analysis of BNN performance under varying hyperparameter settings. 
Our results indicate that many of the hyperparameters interact with each other to affect both predictive accuracy and UQ.
For improved usage of BNNs in real-world applications, we suggest that global sensitivity analysis, or related methods such as Bayesian optimization, should be used to aid in dimensionality reduction and selection of hyperparameters to ensure accurate UQ in BNNs.

\end{abstract}

\section{Introduction}
Robust uncertainty quantification (UQ) is essential when applying machine learning models in high-stakes applications.
For instance, in scientific applications, training a neural network (NN) that can predict well on data similar to that seen during training may not be sufficient, as we would like to know regions of input space where uncertainty increases to potentially collect more data or refine hypotheses.
For instance, UQ can guide experimental design in materials science by identifying high-uncertainty regions in the search space~\cite{tran2020methods}, can help manage risk in statistical downscaling in climate modeling~\cite{vandal2018quantifying}, and can identify when NNs for turbulence modeling~\cite{geneva2019quantifying} or geological analysis~\cite{klein2025bnn} begin to extrapolate to new regimes.
However, while UQ has been essential for application of NNs in scientific areas, \textit{poorly calibrated} UQ can lead to overconfident predictions and erroneous scientific conclusions.
For true integration of machine learning methods such as NNs into science, reliable UQ essential~\cite{abdar2021review}.

Bayesian neural networks (BNNs)~\cite{mackay1992practical,neal1992bayesian} offer a promising approach that combines the empirical success of NN models across a plethora of prediction tasks with the principled UQ that arises from Bayesian modeling~\cite{hoff2009first}. 
However, in practice, BNNs present unique challenges, in part because exact Bayesian inference is not possible (necessitating approximate inference methods) and in part because the complexity of the models and the number of hyperparameters increases relative to standard neural networks (NNs).
In particular, when training BNNs, one must consider prior distribution hyperparameters and hyperparameters specific to the approximate inference method.
In practice, tuning hyperparameters for BNNs appears to be more challenging than for non-Bayesian NNs; what is done in practice often appears closer to an art than a science~\cite{tuningplaybookgithub}. 
This calls into question the reliability of UQ from BNN approaches.

In this paper, we investigate how the performance of BNNs is influenced by hyperparameter choices in two synthetic data sets and seek to offer general guidance for hyperparameter tuning for successful BNN inference.
We focus on the use of variational inference (VI)~\cite{blei2017variational}, a tractable approximate inference method frequently chosen for computationally tractable BNNs~\cite{graves2011practical, li2016renyi}, though we acknowledge that there are alternative inference schemes including Monte Carlo dropout~\cite{gal2016dropout}, deep ensembles~\cite{lakshminarayanan2017simple}, Laplace approximations~\cite{daxberger2021laplace}, and low-rank Gaussian approximations~\cite{maddox2019simple}.
Our methodology for evaluating model performance and investigating sensitivity to hyperparameters could extend to such alternate inference approaches, but in this paper, we focus on general principles and procedures for obtaining good results with VI.


\textit{Good results} are of course subjective, but we focus on quantifying two key aspects of a BNN fit: predictive accuracy of the average prediction and quality of predictive uncertainty intervals. 
In contrast, previous work on methodology for inference for BNNs typically focuses only on error rate or average likelihood.
For instance, the introduction of stochastic variational inference for BNNs focused on phoneme error rate for the TIMIT speech corpus \cite{graves2011practical}, while the expectation of the log likelihood of a test set over the posterior distribution of BNN weights is a popular modern measure of uncertainty for testing novel variational methodologies~\cite{hernandez2015probabilistic, louizos2017multiplicative, louizos2016structured}.  
However, utilizing expected log-likelihood on a test set is only sensible in model comparison contexts and for methodological development, but does not inform a practitioner about whether their predictive uncertainty intervals are reasonable and likely to be useful in a specific application.
In contrast, we assess how hyperparameters affect not only predictive accuracy (via root mean squared error, or RMSE) but also uncertainty intervals (via interval score, or IS, a metric that measures both coverage and interval width), with a focus on extracting information on the role that hyperparameters play in BNN inference. 

Here, we utilize global sensitivity analysis~\citep{andrea2008global} to work towards a more principled process for hyperparameter selection in BNNs. 
Global sensitivity analysis evaluates the impact of hyperparameter choices across an entire range of options and is particularly valuable when the response of interest (in this case, BNN performance) is nonlinear with respect to the hyperparameters. 
Using global sensitivity analysis, we can rank the influence of different hyperparameters and understand how they interact.
In this paper, we evaluate the global sensitivity of two different VI algorithm variants to hyperparameters on two different synthetic data sets.
Our results illustrate the complexity of BNN inference, underscore the necessity of tuning hyperparameters carefully rather than relying on heuristic default values, and open up the possibility for future research to better understand how BNN hyperparameters interact in complex ways to impact the quality of UQ.
We hope our results are useful to those seeking to use BNNs in areas where calibrated UQ is critical, such as in scientific fields as disparate as climate modeling, medicine, physics simulations, and materials design.

In Sec.~\ref{sec:mm}, we give background on BNNs and VI, detail the performance metrics we use, give an overview of global sensitivity analysis, and provide specific experimentation details.  Sec.~\ref{sec:results} summarizes the results of the global sensitivity analysis, including both main effects plots and tabulated sensitivity indices. Finally, in Sec.~\ref{sec:discussion}, we discuss the findings, propose future research related to additional sensitivity analysis investigations, and offer suggestions for practical hyperparameter selection in BNNs.

\section{Materials and Methods} \label{sec:mm}
In this section, we first give background information on BNNs and VI (Secs.~\ref{subsec:bnn} and \ref{subsec:vi}), followed by an exposition of performance metrics we will use to evaluate hyperparameter choices (Sec.~\ref{subsec:metrics}).
Sec.~\ref{subsec:globalsens} explains the global sensitivity analysis we use to determine the impact of hyperparameters on each metric. 
Finally, we discuss our specific experimental procedures in Sec.~\ref{subsec:expmethods}.

\subsection{Bayesian Neural Networks (BNNs)} \label{subsec:bnn}
Traditional NNs propose a function $g(x,\bm{\theta})$ that depends on parameters $\bm{\theta}$ (typically comprised of weights and biases for the neural network layers) and learns to predict outputs $y$ by optimizing over $\bm{\theta}$ using training pairs $\{(x_i, y_i)\}_{i=1}^n$ and a loss function that seeks to align $g$ with the observed $y$ values.
In contrast, Bayes' rule posits a posterior distribution over parameters rather than a single setting of parameters: 
\begin{equation*}
p(\bm{\theta}|x,y) = \frac{p(y|x,\bm{\theta})p(\bm{\theta})}{p(y|x)} \propto p(y|x,\bm{\theta})p(\bm{\theta}).
\end{equation*}
In Bayesian inference, the goal is to infer the posterior distribution \(p(\bm{\theta}|x,y)\) of our model parameters \(\bm{\theta}\) given our data observations $\{(x_i, y_i)\}_{i=1}^n$ using a likelihood function \(p(y|x,\bm{\theta})\) and prior distribution \(p(\bm{\theta})\)~\cite{mackay1992practical,neal1992bayesian,goan2020bayesian}.  
In BNNs, the likelihood is typically specified by the NN function $f$ as part of some chosen conditional probability distribution; for instance, for regression tasks, a common likelihood is Gaussian: $p(y|x,\bm{\theta})\sim \mathcal{N}(g(x,\bm{\theta}),\sigma^2)$ where $\sigma^2$ is a noise variance that can be fixed or learned.
Prior distributions are often specified with simple distributions such as independent standard Gaussians for each weight~\cite{jospin2022hands}, though recent works have explored alternate choices~\cite{fortuinbayesian,tran2022all}.
Given the prior and likelihood, exact evaluation of the posterior distribution is typically not possible for complicated models such as neural networks.
Markov Chain Monte Carlo (MCMC, \cite{RobertChristianP.2004MCsm}) is a common method to produce samples from the posterior distribution, but presents challenges in neural networks due to high dimensionality and multimodality of the parameter space~\cite{papamarkou2022challenges}.
Therefore, most existing applications of Bayesian neural networks rely on approximate inference schemes such as variational inference, discussed in the next section.

Given an approximation of the posterior distribution, predictive uncertainty in BNNs is obtained via the posterior predictive distribution
\begin{align*}
    p(y^* \, | \, x^*, x, y) = \int p(y^* \, | \, x^*, \bm{\theta}, \sigma^2) p(\theta  \, | \, x, y) d\theta d\sigma^2. 
\end{align*}
That is, for a new input $x^*$, we can compute the probability of the output $y^*$ conditional on $x^*$ by integrating over parameters sampled from the posterior.
Note that in this formulation we have treated $\sigma^2$ as a known and fixed noise variance as part of the likelihood.
In practice, we will approximate this integral with samples, so a typical procedure would be to repeatedly sample $\bm{\theta}$ from the approximate posterior, then sample $y^*$ from the likelihood given both $\bm{\theta}$ and the fixed noise variance $\sigma^2$.

\subsection{Variational Inference (VI)} \label{subsec:vi}
Because exact posterior inference for BNNs is difficult, typical BNN methods approximate the posterior distribution. 
One approximation method, variational inference (VI), aims to minimize a statistical distance from a variational approximate posterior \(q(\bm{\theta})\) chosen to lie in some distributional family to the true posterior distribution \(p(\bm{\theta}|x,y)\)~\cite{blei2017variational}.  
In this work, we compare results using two different measures of statistical distance. 
The first, Kullback Leibler (KL) divergence~\cite{kullback1951information}, is more commonly used in BNNS~\cite{graves2011practical,blundell2015weight} and its use can be considered analogous to viewing Bayesian inference as an optimization problem~\cite{knoblauch2022optimization} when \(q()\) is the same distributional form as the posterior.
We utilize the equality
\begin{equation*}
\mathrm{KL}[q(\bm{\theta})||p(\bm{\theta}|x,y)] =  - \mathbb{E}_{q(\bm{\theta})}\left[p(y|x,\bm{\theta})\right] + \mathrm{KL}[q(\bm{\theta})||p(\bm{\theta})] + \log(p(y|x))
\end{equation*}
in the estimation; because for a fixed set of training data and prior, \(\log(p(y|x))\) is constant and positive, we have that
\begin{equation}
\label{eq:klbound}
\mathrm{KL}[q(\bm{\theta})||p(\bm{\theta}|x,y)] \le  - \mathbb{E}_{q(\bm{\theta})}\left[p(y|x,\bm{\theta})\right] + \mathrm{KL}[q(\bm{\theta})||p(\bm{\theta})].
\end{equation}
The right hand side forms and objective function that we can minimize in an effort to minimize the KL divergence on the left hand side by bounding it from above.
For typical choices of prior and $q$, the KL divergence in Eq.~\ref{eq:klbound} is available in closed form, but the expectation must generally be estimated via Monte Carlo sampling.
In practice, the objective function that is utilized is often written
\begin{equation}
    \label{eq:klobjective}
    \mathcal{L}_{\mathrm{KL}}(q) = - \mathbb{E}_{q(\bm{\theta})}\left[p(y|x,\bm{\theta})\right] + \gamma \mathrm{KL}[q(\bm{\theta})||p(\bm{\theta})].
\end{equation}
The parameter \(\gamma\) on the right hand side of Equation \ref{eq:klobjective} is naturally set equal to one for the inequality to hold. 
However, in practice, values of \(\gamma\) are chosen to downweight or upweight the influence of the prior distribution on inference~\cite{wenzel2020good}.
In addition to $\gamma$, other hyperparameters for variational inference with $\mathcal{L}_\mathrm{KL}$ include the prior hyperparameters, the number of Monte Carlo samples used for estimating the expectation, the learning rate for optimization, and the number of steps in the optimization.

Another choice of divergence that has been recently explored is $\alpha$-Renyi divergence~\cite{li2016renyi,knoblauch2022optimization}.
The divergence is given by
\begin{equation}
\label{eq:arenyidivergence}
\mathcal{L}_{AR}(q) = D_{\mathrm{AR}}^{\alpha}[q(\theta) \, || \, p(\theta|y)] = \frac{1}{\alpha -1} \log \left(\int p(\theta|y)^{\alpha}q(\theta)^{1-\alpha} d\theta\right).
\end{equation}
We note that for $\alpha \rightarrow 1$, one can show that this divergence becomes the KL divergence, so it can be seen as a generalization of KL divergence.
The hyperparameters are similar to those for $\mathcal{L}_\mathrm{KL}$ with the addition of the $\alpha$ parameter, but there is no $\gamma$ weighting term. 
Similar to KL divergence, the integral is evaluated via Monte Carlo sampling.

\subsection{BNN Performance Metrics} \label{subsec:metrics}
We investigate the sensitivity two performance metrics for quantifying the goodness of model fit to settings of our VI hyperparameters. 
Often, the performance of NNs and BNNs is judged using a measure of accuracy such as root mean squared error (RMSE; Eq.~\ref{eq:rmse}) comparing the values of a testing set to the expected function value \(\hat{g}(x_i)\) predicted from the fitted model.
For BNNs, it is common to use the mean of the posterior predictive distribution to make predictions $\hat{g}$ for accuracy computations.
The ideal RMSE value for a model is a function of the true noise of the data generating mechanism as well as the number of test points and the uniformity of the test points in the input space \(X\), but a lower RMSE is typically considered a better fit. 
RMSE is defined as
\begin{equation}
    \label{eq:rmse}
    \mathrm{RMSE} = \sqrt{\frac{1}{N} \sum_{i = 1}^{N_{\mathrm{test}}} [y_{\mathrm{test}}(x_i) - \hat{g}(x_i)]^2}.
\end{equation}

However, RMSE does not directly inform about predictive UQ, so we additionally compute the interval score (IS; Eq.~\ref{eq:is}) \citep{hyndman1996computing,gneiting2007strictly} on our testing set based on samples from the posterior predictive distribution.
Intuitively, the IS measures the accuracy of a prediction interval produced at the \(\alpha_{\mathrm{IS}}\) level with an upper bound of \(u(x^{\mathrm{IS}}_i)\) and a lower bound of \(l(x^{\mathrm{test}}_i)\) for given testing input location \(x^{\mathrm{test}}_i\) and is defined as
\begin{equation}
\label{eq:is}
\begin{split}
\mathrm{IS} &= \sum_{i = 1}^{N_{\mathrm{test}}} (u(x^{\mathrm{test}}_i) -l(x^{\mathrm{test}}_i)) + \frac{2}{\alpha_{\mathrm{IS}}}(l(x^{\mathrm{test}}_i) - y_{\mathrm{test}}(x^{\mathrm{test}}_i)) {1}(y_{\mathrm{test}}(x^{\mathrm{test}}_i) < l(x^{\mathrm{test}}_i))\\
&+\frac{2}{\alpha_{\mathrm{IS}}}(y_{\mathrm{test}}(x^{\mathrm{test}}_i) - u(x^{\mathrm{test}}_i) ) {1}(y_{\mathrm{test}}(x^{\mathrm{test}}_i) > u(x^{\mathrm{test}}_i)).
\end{split}
\end{equation}
The function \({1}(a < b)\) is equal to 1 when the equality contained within the parenthesis is true, and equal to 0 when the equality is false.
A lower IS is considered to indicate a better fit. The intuition is that we want predictive intervals that contain the true values, but that are not overly wide.
The first term in Eq.~\ref{eq:is} penalizes intervals which are too wide. 
The second and third terms penalize testing data points which lie outside of the prediction interval.  



\subsection{Global Sensitivity Analysis} \label{subsec:globalsens}

Our goal is to investigate how varying hyperparameter settings, and combinations of hyperparameter settings, affects RMSE and IS in BNNs.  
Global sensitivity analysis provides such results \citep{andrea2008global}. 
Specifically, we will investigate how much a single factor can independently affect RMSE and IS through \textit{main effects} and \textit{first order sensitivity indices}, as well as how the interaction of a parameter with other parameters affects results through \textit{total sensitivity indices}.  
We choose global sensitivity analysis for our investigation because of the limitations of local derivative-based sensitivity for functions of uncertain form (see Chapter 1 of~\citep{andrea2008global}). 
Essentially, when a derivative is evaluated under function uncertainty and the function is not first-order linear, interpretation becomes difficult both at the location of evaluation as well as when extrapolating to a larger domain.  
Global sensitivity analysis instead considers a set of parameter ranges and weights them under the consideration of a measure on their probability of occurrence. 
This allows us to examine how factors influence the outcome both individually and in combination, crucial for nonlinear responses with correlated inputs. 
In this case, we consider marginally uniform distributions via Latin hypercube sampling (LHS), but other \textit{prior} distributional forms of the parameters can be considered \citep{saltelli2002relative, gramacy2008bayesian}. 

In global sensitivity analysis, we typically sample hyperparameters $\omega$ and evaluate the response property of interest (for instance, the BNN RMSE) for each hyperparamter setting. 
Because this process is time-consuming (requiring fitting a BNN for each selection of hyperparameters), we utilize a surrogate model $f(\omega)$ that approximates the BNN performance for a new setting of hyperparameters (Sec.~\ref{subsec:tgp}). Based on this surrogate model, we can investigate a number of metrics of sensitivity, including main effects (Sec.~\ref{subsec:maineff}), first order sensitivity indices (Sec.~\ref{subsec:foeff}), and total sensitivity indices (Sec.~\ref{subsec:toteff}) while incorporating uncertainty in the surrogate model into the analysis.

\subsubsection{Treed Gaussian Process Surrogate Model} \label{subsec:tgp}
While a variety of surrogate models could be utilized for global sensitivity analysis, here we use a fully Bayesian surrogate model to incorporate uncertainty in the surrogate model into our analysis.
Because we do not expect the response to be smooth with respect to the hyperparameters, we use a treed Gaussian process (TGP) \cite{gramacy2008bayesian} for each response of interest (RMSE and IS).
TGPs are flexible models frequently used to emulate responses and partition the input space with separately parameterized Gaussian process (GP) models in order to allow for further flexibility. 
Specifically, we propose a surrogate 
\begin{align*}
    f(\omega) \sim GP_r(\mu_r(\omega),k_r(\omega,\omega') \text{ for } \omega \text{ in region } r,
\end{align*}
where $\mu_r$ is the local Gaussian process mean function for region $r$, $k_r$ is the covariance function for region $r$, and the regions are determined using a tree structure that splits the input space.
We utilize the \texttt{tgp} package \citep{gramacy2010categorical} in \textsf{R}, which provides a fully Bayesian implementation of the model using Markov Chain Monte Carlo (MCMC) to sample from the posterior distribution.
The \texttt{tgp} package additionally provides functionality for global sensitivity analysis \citep{andrea2008global}, leveraging the TGP model fit to a provided set of data to approximate the integrals required for global sensitivity analysis.   
Further details of implementation can be found in Sec.~\ref{subsec:code} and \cite{gramacy2010categorical}. 

\subsubsection{Main Effects} \label{subsec:maineff}
Main effects can be considered as the effect of a single parameter of interest, if we average, or take the expectation, over the other parameters:
\begin{equation*} 
\mathbb{E}_{\omega_{-j}}[f(\omega) | \omega_{j}] = \int f(\omega) \ p(\omega_{-j}|\omega_{j}) d\omega_{-j}.
\end{equation*}
Here, \(\omega_j\) is our \(j^{\mathrm{th}}\) parameter of interest for the main effect, and \(\omega_{-j}\) is the set of parameters excluding parameter \(\omega_j\), where we assume a sufficiently large LHS sample such that \(p(\omega_{-j}|\omega_j) \approx p(\omega_{-j})\).
The result of the numerical evaluation of this integral is a function which can be evaluated for a value of \(\omega_j\) when averaging over possible function evaluations of the remaining \(\omega_{-j}\) parameters. 
That is, the main effect gives us the expected value of $f$ for a given setting of $\omega_j$ while averaging across all other hyperparameters.
With a fully Bayesian surrogate model such as a TGP, we also obtain uncertainty of the main effect.
Uncertainty quantification of the main effect is related to GP model uncertainty \cite{gramacy2010categorical}.  The main effect is numerically approximated conditioned on a set of sampled GP parameters, and the process is repeated within each iteration of an MCMC algorithm (drawing from the posterior distribution over $f$) for generating posterior draws of model parameters. 

\subsubsection{First Order Sensitivity Indices} \label{subsec:foeff}
The first order sensitivity index is a measurement of how a single factor \(\omega_j\) contributes to the variance of the outcome of a function, under the assumption that the remaining factors are held constant.  If we think of the main effect \(\mathbb{E}_{\omega_{-j}}[f(\omega)|\omega_j]\) as a function of variable \(\omega_j\), then the first order sensitivity index is related to the variance, with respect to \(\omega_j\), of the main effect function as \(\int\left(\mathbb{E}_{\omega_{-j}}[f(\omega)|\omega_j] - \mathbb{E}_{\omega}[f(\omega)]\right)^2 p(\omega_j) d\omega_j\).

The sensitivity analysis functions in the \texttt{tgp} \textsf{R} package only consider the variance of the surrogate function, and ignores the noise of the response for sensitivity analysis \cite{tgppkg}. Therefore, the first order sensitivity index is the fraction of the total variance of our surrogate function which is explained by input \(\omega_j\):
\begin{equation}\label{eq:firstorder}
S_j = \frac{\mathbb{V}\mathrm{ar}_{\omega_j}(\mathbb{E}_{\omega_{-j}}[f(\omega)|\omega_j])}{\mathbb{V}\mathrm{ar}(f(\omega))}.
\end{equation}

Another way to consider the interpretation of the first order sensitivity analysis relies on the law of total variance decomposition \(\mathbb{V}\mathrm{ar}(f(\omega)) = \mathbb{E}_{\omega_j}[\mathbb{V}\mathrm{ar}_{\omega_{-j}}(f(\omega)|\omega_j)] + \mathbb{V}\mathrm{ar}_{\omega_j}(\mathbb{E}_{\omega_{-j}}[f(\omega)|\omega_j])\).  Dividing both sides by \(\mathbb{V}\mathrm{ar}(f(\omega))\) leads to
\begin{equation*}
\frac{\mathbb{V}\mathrm{ar}(f(\omega))}{\mathbb{V}\mathrm{ar}(f(\omega))} = \frac{\mathbb{E}_{\omega_j}[\mathbb{V}\mathrm{ar}_{\omega_{-j}}(f(\omega)|\omega_j)]}{\mathbb{V}\mathrm{ar}(f(\omega))} + \frac{\mathbb{V}\mathrm{ar}_{\omega_j}(\mathbb{E}_{\omega_{-j}}[f(\omega)|\omega_j])}{\mathbb{V}\mathrm{ar}(f(\omega))}.
\end{equation*}

In the right hand summand, \(\mathbb{V}\mathrm{ar}_{\omega_j}(\mathbb{E}_{\omega_{-j}}[f(\omega)|\omega_j])\) is the variance of our function \(f(\omega)\) if we only chose to model \(\omega_j\), the variance of our main effect, while \(\mathbb{E}_{\omega_j}[\mathbb{V}\mathrm{ar}_{\omega_{-j}}(f(\omega)|\omega_j)]\) is the residual variation in \(f(\omega)\) that can be attributed to the remaining \(\omega_{-j}\) parameters and interactions between \(\omega_j\) and \(\omega_{-j}\). Dividing each term by \(\mathbb{V}\mathrm{ar}(f(\omega))\) produces the fraction of the variation of \(f(\omega)\).
Because \(\mathbb{E}_{\omega_j}[\mathbb{V}\mathrm{ar}_{\omega_{-j}}(f(\omega)|\omega_j)]\) is the residual variation this term contains the first order sensitivity of each \(\omega_{-j}\), as well as terms related to the total effect of \(\omega_j\).  Next, we will show how to decouple the total effect terms for an \(\omega_j\) from the residual variance \(\mathbb{E}_{\omega_j}[\mathbb{V}\mathrm{ar}_{\omega_{-j}}(f(\omega)|\omega_j)]\).

\subsubsection{Total Sensitivity Indices} \label{subsec:toteff}
A total sensitivity index, \(T_i\) of factor \(\omega_j\) is the sum of all first order and higher sensitivity index terms related to factor \(\omega_j\).  The law of total variance allows for the decomposition
\begin{equation}\label{eq:totalfull}
 \mathbb{V}\mathrm{ar}_{\omega_j}(\mathbb{E}_{\omega_{-j}}[f(\omega)|\omega_j]) +  \mathbb{E}_{\omega_j}[\mathbb{V}\mathrm{ar}_{\omega_{-j}}(f(\omega)|\omega_j)] =  \mathbb{V}\mathrm{ar}_{\omega_{-j}}(\mathbb{E}_{\omega_{j}}[f(\omega)|\omega_{-j}]) + \mathbb{E}_{\omega_{-j}}[\mathbb{V}\mathrm{ar}_{\omega_{j}}(f(\omega)|\omega_{-j})].
\end{equation}

\sloppy On the left hand side of Eq.~\ref{eq:totalfull}, we have terms related to the first order sensitivity (\( \mathbb{V}\mathrm{ar}_{\omega_j}(\mathbb{E}_{\omega_{-j}}[f(\omega)|\omega_j])\)) and the remainder of the total variance (\(\mathbb{E}_{\omega_j}[\mathbb{V}\mathrm{ar}_{\omega_{-j}}(f(\omega)|\omega_j)]\)). 
On the right hand side, we have $\mathbb{V}\mathrm{ar}_{\omega_{-j}}(\mathbb{E}_{\omega_{j}}[f(\omega)|\omega_{-j}])$, which is related to the all first order and higher sensitivity indices for all \(\omega_{-j}\) factors \cite{andrea2008global}.  The term \(\mathbb{E}_{\omega_{-j}}[\mathbb{V}\mathrm{ar}_{\omega_{j}}(f(\omega)|\omega_{-j})]\) includes variability related to \(\omega_j\) and interactions between \(\omega_j\) and \(\omega_{-j}\).  We can therefore find the total sensitivity indices as
\begin{equation}\label{eq:secondorder}
T_j = S_j + \frac{\mathbb{E}_{\omega_j}[\mathbb{V}\mathrm{ar}_{\omega_{-j}}(f(\omega)|\omega_j)] - \mathbb{V}\mathrm{ar}_{\omega_{-j}}(\mathbb{E}_{\omega_{j}}[f(\omega)|\omega_{-j}])}{\mathbb{V}\mathrm{ar}(f(\omega))} = 1 - \frac{\mathbb{V}\mathrm{ar}_{\omega_{-j}}(\mathbb{E}_{\omega_{j}}[f(\omega)|\omega_{-j}])}{\mathbb{V}\mathrm{ar}(f(\omega))}.
\end{equation}
The term \(\mathbb{E}_{\omega_j}[\mathbb{V}\mathrm{ar}_{\omega_{-j}}(f(\omega)|\omega_j)] - \mathbb{V}\mathrm{ar}_{\omega_{-j}}(\mathbb{E}_{\omega_{j}}[f(\omega)|\omega_{-j}])\) removes explanations of variance irrelevant to \(\omega_j\) from the residual variance term found in Eq.~\ref{eq:firstorder} and summing with \(S_j\) produces the total sensitivity index.
If we considered \(\omega_{-j}\) to be a vector instead of a scalar parameter, we could say \(\mathbb{V}\mathrm{ar}_{\omega_{-j}}(\mathbb{E}_{\omega_{j}}[f(\omega)|\omega_{-j}])\) is the equivalent to the \textit{first order index} of vector parameter \(\omega_{-j}\).  
The right most side of the equality in Eq.~\ref{eq:secondorder} shows removal of the proportion of variance explained by \(\omega_{-j}\) from the total variance of \(f(\omega)\).
In interpreting the total sensitivity indices, it is often useful to compare them to the first order sensitivity indices; the difference between the two gives information on how much interactions with other parameters drive the total effect of a given parameter.

\subsection{Experimental Methods} \label{subsec:expmethods}
In this section, we first describe synthetic data generating mechanisms we used as test cases for fitting BNNs (Sec.~\ref{subsec:datagen}); we then give details on our BNN implementations (Sec.~\ref{subsec:bnnmodeling}), our experimental setup (Sec.~\ref{subsec:expsetup}), and the code to reproduce our results (Sec.~\ref{subsec:code}).

\subsubsection{Data Generating Mechanisms} \label{subsec:datagen}
For the global sensitivity analysis, we chose two simple data generating mechanisms that differ in the number of inputs to the underlying function; both data generating mechanisms have continuous inputs and outputs and homoskedastic noise. 
We avoided difficult test functions to ensure the difficulty of the problem does not confound the results (that is, ensuring the BNNS can fit the underlying function reasonably well without needing thorough architecture search). 
The first data generating mechanism is a polynomial function of one input:
\begin{align}\label{eq:x1d}
y = x^3 - x^2 + 3 + \epsilon, \quad \epsilon \sim \mathcal{N}(0, \sigma = 0.5).
\end{align}
This data generating mechanism is similar to what is used as a demo function for the \texttt{torchbnn} package \cite{lee2022graddiv}.  Because the input dimension of Eq.~\ref{eq:x1d} is \(d = 1\), we refer to this data generating mechanism as \(X \in \mathbb{R}^1\) in shorthand.
Training and testing data were obtained as uniformly random on \(X \in [-2, 2]\). 

\begin{figure}
\centering
\includegraphics[width=\linewidth]{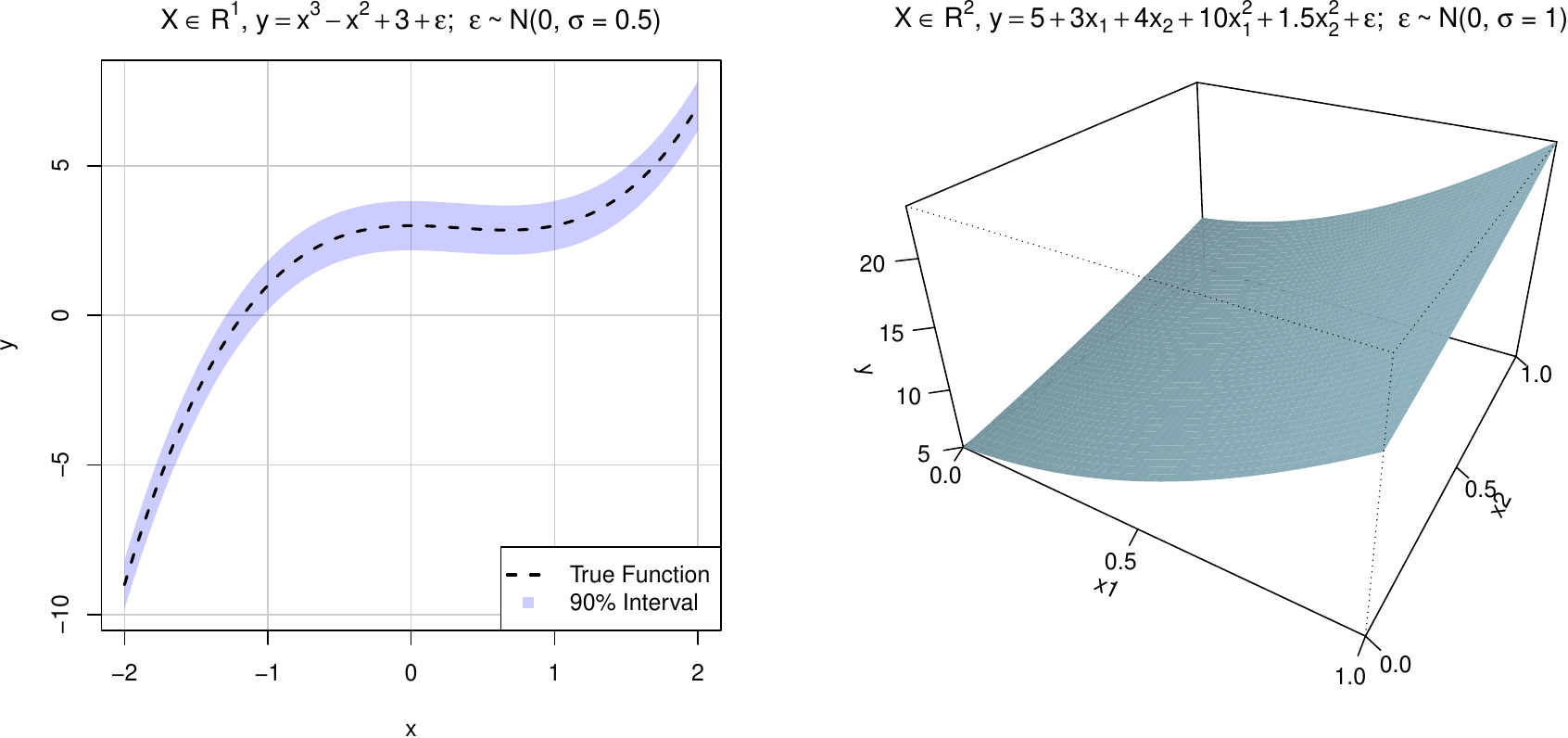}
\caption{The data generating mechanisms used for testing the variational BNN approximations.  For the \(X \in \mathbb{R}^1\) data generating mechanism (left) the expected value of the function is plotted along with the 90\% conditional distribution interval arising from additive homoskedastic noise.  For the \(X \in \mathbb{R}^2\) data generating mechanism (right), we show the expected value of the function with respect to the two input dimensions.}
\label{fig:dgm}
\end{figure}

The second data generating mechanism we investigate is
\begin{align}\label{eq:x2d}
y = 5 + 3 x_1 + 4 x_2 + 10x_1^2 + 1.5 x_2^2 + \epsilon, \quad \epsilon \sim \mathcal{N}(0, \sigma = 1).
\end{align}
We refer to the data generating mechanism in Eq.~\ref{eq:x2d} as \(X \in \mathbb{R}^2\) because the dimensionality of the input space is \(d = 2\).
For \(X \in \mathbb{R}^2\) training and testing data was uniformly sampled on \([0,1]^2\). 
Plots of the data generating mechanisms are shown in Fig.~\ref{fig:dgm}.

\subsubsection{BNN Implementation} \label{subsec:bnnmodeling}
For our BNNs, we assume each observation is independently and identically distributed, given the known noise variance \(\sigma^2\), a set of neural network weights and biases \(\bm{\theta}\), and location \(x_i\).  
We choose a univariate Gaussian likelihood function, with the mean specified as the output of the NN $g(x,\bm{\theta})$ and known noise variance $\sigma^2$.  
Each neural network weight and bias in the set of NN parameters \(\bm{\theta}\) has its own independent univariate normal prior distribution with identical prior means \(\mu_0\) and prior variances \(\sigma_0^2\) for each parameter. 
We note that both \(\mu_0\) and \(\sigma_0^2\) are varied as part of our sensitivity analysis.

The NN architectures we use consist of two layers, with the output of the first layer passed through a \(\mathrm{tanh}(x)\) activation function.  The number of features passed between the NN layers is varied in our experiments and ranges from 2 to 100.  A bias vector is added to the output of the first layer and a bias scalar value is added to the output of the second layer.  For the \(X \in \mathbb{R}^1\) mechanism, the inputs for both testing and training data are scaled to \([0,1]\); rescaling inputs is not necessary for the \(X \in \mathbb{R}^2\) mechanism.  We assume in our setup that the noise variance is known so that our investigation focuses on the variance of the BNN functional output.  For both the \(X\in \mathbb{R}^1\) and \(X\in \mathbb{R}^2\) mechanisms we rescale training responses \(y_{\mathrm{train}}\) and testing responses \(y_{\mathrm{test}}\) as 
\begin{equation*}
\frac{y - \frac{1}{N_{\mathrm{train}}}\sum_{i = 1}^{N_{\mathrm{train}}}h(x_i^{\mathrm{train}})}{\sqrt{\sigma^2 + \frac{1}{N_{\mathrm{train}}}\sum_{i = 1}^{N_{\mathrm{train}}}\left(h(x_i^{\mathrm{train}}) - \frac{1}{N_{\mathrm{train}}}\sum_{i = 1}^{N_{\mathrm{train}}}h(x_i^{\mathrm{train}})\right)^2}}.
\end{equation*}
Where, \(h(x_i^{\mathrm{train}})\) is the data generating mechanism evaluated at the \(i^{\mathrm{th}}\) training data location \(x_i^{\mathrm{train}}\) before noise is added, \(N_{\mathrm{train}}\) is the number of training data points, and \(\sigma^2\) is the known noise variance.  

For our variational approximations, we choose independent Gaussian distributions \(q(\theta_i|\hat{\mu}_i, \hat{\sigma}^2_i)\) for weight or bias \(\theta_i\) which has a uniquely estimated mean \(\hat{\mu}_i\) and variance \(\hat{\sigma}_i^2\). 
Therefore, the number of parameters required for estimation through optimization is twice the number of non-Bayesian NN weights and biases.
For fitting, we use stochastic gradient descent and estimate any expectations in the loss functions via Monte Carlo sampling with the reparameterization trick (Bayes by Backprop; see~\cite{blundell2015weight}). 
The learning rate, number of optimizer steps, and number of Monte Carlo samples are all varied in the sensitivity analysis. 

\subsubsection{Experimental Setup} \label{subsec:expsetup}
We collected data to analyze how the hyperparameter choices required for variational approximations affect the RMSE and IS metrics.  
The parameters we vary are the KL reweighting \(\gamma\) (Eq.~\ref{eq:klobjective}), the \(\alpha\) parameter utilized in calculating \(\alpha -\)Renyi divergence (Eq.~\ref{eq:arenyidivergence}), the prior mean $\mu_0$, the prior standard deviation \(\sigma_0\), the number of optimizer steps, the number of NN features, the number of samples drawn from the variational posterior in order to approximate the integrals required for KL divergence or \(\alpha - \)Renyi divergence, and the learning rate (LR).  
Instead of uniformly sampling over the natural values of \(\gamma, \sigma_0, \) and LR, we uniformly sample over \(\log_{10}\) of these values in order to better space out the smaller values.  

To obtain a well spaced set of data for our parameters of interest, we generated a 750-point 7-dimensional Latin hypercube sample (LHS)~\citep{mckay2000comparison} across the hyperparameters in Table~\ref{tab:expparams} for each of the four combinations of data generating mechanism ($\mathbb{R}^1$ or $\mathbb{R}^2$) and statistical distance (KL or $\alpha$-Renyi). 
The points in an LHS are uniformly distributed marginally over each of the 7 dimensions, but have constraints which ensure the points are well spaced within the full 7 dimensional space. 
Our LHS was bounded by a hyper-rectangle with bounds shown in Table~\ref{tab:expparams}.
The bounds are the same for both data generating mechanisms and choices of statistical distance.  
For each hyperparameter setting, we fit a BNN to generated data and evaluated the RMSE and IS on a held-out test set for sensitivity analysis.

\begin{table}[!h]
\centering
\caption{\label{tab:expparams}Bounds for hyperparameters explored with global sensitivity analysis.  For a given statistical distance (KL or $\alpha$-Renyi), the bounds shown are used with both the $X \in \mathbb{R}^1$ and $X \in \mathbb{R}^2$ data generating mechanisms.}
\centering
\begin{tabular}[t]{lrrrrrrrr}
\toprule
    & $\log_{10}(\gamma)$ & $\alpha$ & $\log_{10}(\sigma_0)$ & \makecell{Optimizer\\Steps} & NN Features & \makecell{Samples for\\Integration} & $\log_{10}$(LR) & $\mu_{0}$\\
\midrule
\addlinespace[0.3em]
\multicolumn{9}{l}{\textbf{KL Parameters}}\\
\hspace{1em}\cellcolor{gray!10}{Upper Bound} & \cellcolor{gray!10}{1} & \cellcolor{gray!10}{NA} & \cellcolor{gray!10}{0.5} & \cellcolor{gray!10}{20000} & \cellcolor{gray!10}{100} & \cellcolor{gray!10}{25} & \cellcolor{gray!10}{-0.3} & \cellcolor{gray!10}{2}\\
\hspace{1em}Lower Bound & -1 & NA & -0.5 & 2000 & 2 & 1 & -3.3 & -2\\
\addlinespace[0.3em]
\multicolumn{9}{l}{\textbf{$\alpha -$Renyi Parameters}}\\
\hspace{1em}\cellcolor{gray!10}{Upper Bound} & \cellcolor{gray!10}{NA} & \cellcolor{gray!10}{1} & \cellcolor{gray!10}{0.5} & \cellcolor{gray!10}{20000} & \cellcolor{gray!10}{100} & \cellcolor{gray!10}{25} & \cellcolor{gray!10}{-0.3} & \cellcolor{gray!10}{2}\\
\hspace{1em}Lower Bound & NA & 0 & -0.5 & 2000 & 2 & 1 & -3.3 & -2\\
\bottomrule
\end{tabular}
\end{table}

\subsubsection{Code and Software} \label{subsec:code}

Our code for generating data and analysis can be found at the GitHub repository \url{https://github.com/lanl/multiverse/tree/main} \cite{Klein_multiverse_2023} in the \texttt{sensitivity\_bnns} folder.  
Within the \texttt{src} folder the functions \texttt{kl\_div.py} and \texttt{alpha\_renyi.py} define functions which leverage the \texttt{torchbnn} \cite{lee2022graddiv} and \texttt{PyTorch} \cite{paszke2019pytorch} packages to fit variational BNNs with KL divergence and \(\alpha -\)Renyi divergence respectively.  \texttt{evaluate.py} provides functions for obtaining predictions and uncertainty quantification required for evaluating RMSE and IS.  \texttt{prepare\_experiment.py} generates the testing and training data used for each BNN fit as well as the LHS sample specifying tuning parameter choices varied for the experiment.  \texttt{main.py} is run for each row of the LHS defining the experiment to fit a BNN, quantify its performance, and save the results.  \texttt{merge\_results.py} helps to merge results obtained in parallel into one \texttt{*.csv} file for sensitivity analysis using \textsf{R} code.  Details on additional script functionality are provided within the \texttt{sensitivity\_bnns} folder specific \texttt{README.md} file.

Our sensitivity analysis utilized the \texttt{sens()} function within the \texttt{tgp} package.  
We chose to model RMSE and IS with a Bayesian treed Gaussian process, setting the argument \texttt{model = btgp}.  
A first order linear model is specified as the prior mean of the GP. 
The \texttt{sens()} function utilizes MCMC algorithms to obtain posterior draws of the model parameters which are then iteratively utilized with numerical methods to calculate main effects, first order sensitivity indices, and total sensitivity indices\cite{gramacy2010categorical}.  
For MCMC sampling of posterior parameters we specify the argument \texttt{BTE = c(5000, 150000,10)} which stipulates that a total of 150,000 posterior samples are obtained, the first 5000 are discarded under the assumption that the algorithm did not yet converge on the target distribution, and only every \(10^{\mathrm{th}}\) sample was utilized to calculate sensitivity statistics to reduce effects of autocorrelation \cite{RobertChristianP.2004MCsm}.  
We set the argument \texttt{nn.lhs = 7000} for our numerical methods approximating the integrals over the input space required to obtain the desired sensitivity statistics.  
This setting results in 63,000 predictive location evaluations for the Monte Carlo integration scheme for each GP posterior parameter sample considered.  
The predictive locations were bounded by the parameter ranges which can be found in Table~\ref{tab:expparams}.  
The remainder of the function settings were the default settings outlined in the package documentation \cite{tgppkg}.

\section{Results}\label{sec:results}
\begin{figure}
\centering
\begin{adjustbox}{width=1.05\linewidth, center}
\includegraphics[trim=0.1cm 0.05cm 0.05cm 0.1cm, clip]{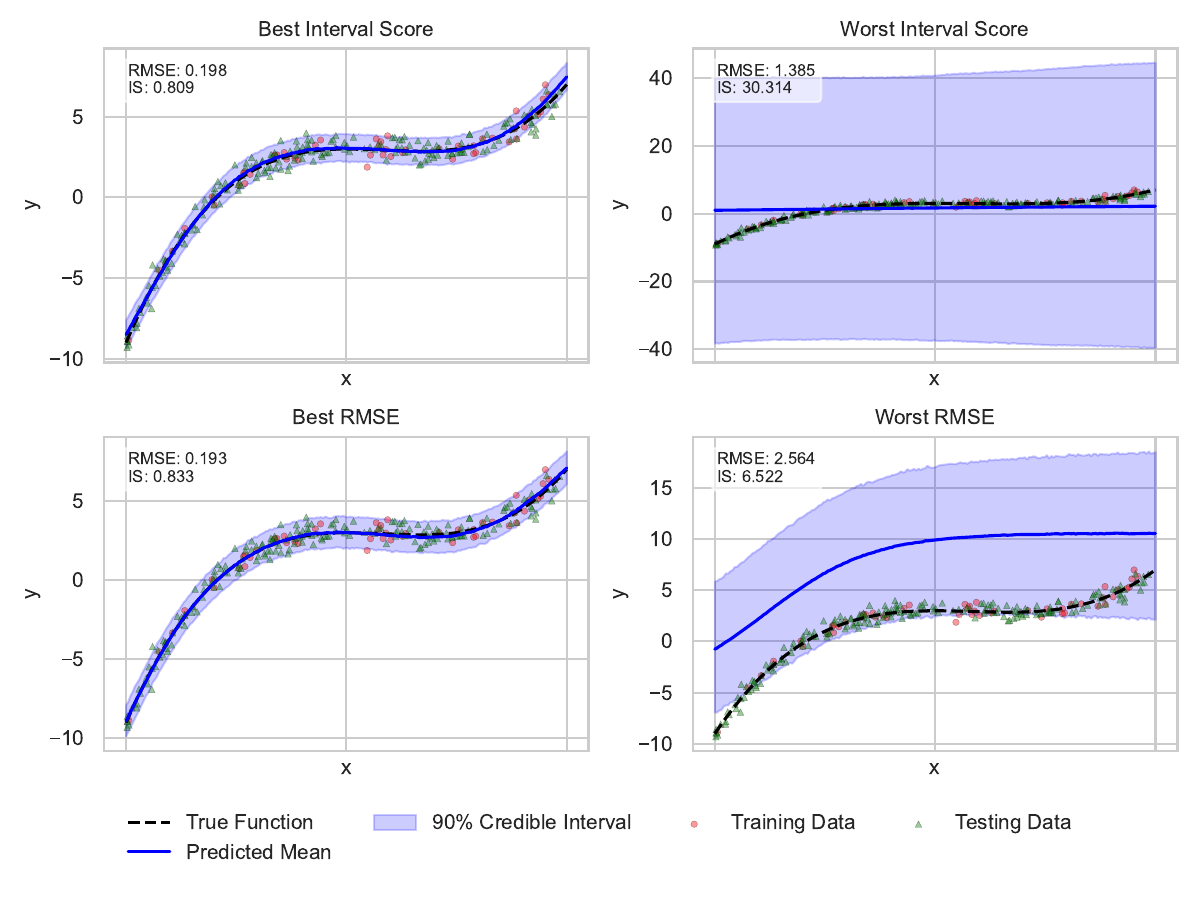}
\end{adjustbox}
\caption{BNN variational approximations through minimizing KL divergence.  The best and worst resulting fits from the 750 different initializations used to collect data are shown. The best fits were selected as either the minimum observed RMSE or IS, and the worst were selected as either the largest observed RMSE or IS. The best fits according to interval score and RMSE are qualitatively similar and appear to describe the data distribution accurately. The worst fit according to interval score appears flat with very high variance, while the worst fit according to RMSE incurs high bias and moderately high variance.} \label{fig:bnnfit}
\end{figure}

In this section, we first give high-level sensitivity analysis results based on first order and total sensitivity indices, then proceed to discuss the main effects of each hyperparameter in detail.
In Fig.~\ref{fig:bnnfit}, we show examples of the best and worst fits obtained for KL divergence with respect to both interval score and RMSE for the $X \in \mathbb{R}^1$ data generating mechanism. 
Qualitatively, the best fits are similar and do a good job capturing the conditional distribution of the data, indicating that well-fitting BNNs can achieve good performance with respect to both metrics.
The worst-case fits, on the other hand, show evidence of both bias and high variance, leading to poor RMSE and interval scores.
The hyperparameters corresponding to each fit are given in Table~\ref{tab:bestworst}.

Table~\ref{tab:kltable} gives the first order and total sensitivity indices for BNNs when KL divergence is used as the loss function, while Table~\ref{tab:arenyitable} gives the indices for BNNs when $\alpha$-Renyi divergence is used as the loss function.
In both cases and across both data generating mechanisms, the learning rate (LR) has the largest first order and total effects, indicating that selection of a good learning rate is essential in BNN fitting. 
For KL divergence, the first order sensitivity indices of \(\log_{10}(\gamma)\) and optimizer steps are larger for \(X \in \mathbb{R}^1\) compared to \(X \in \mathbb{R}^2\), suggesting that these hyperparameters need to be tuned for each specific data set. 
First order sensitivity indices for \(\alpha\)-Renyi divergence are relatively lower than for KL divergence, suggesting that changing specific hyperparameters in isolation may have less effect on VI with $\alpha$-Renyi divergence than with KL divergence.

\begin{table}
\centering
\caption{\label{tab:kltable} First order and total sensitivity indices, displayed as first order (total), for each tuning parameter when KL divergence is used for variational inference; top section corresponds to the RMSE metric sensitivity while bottom section corresponds to the interval score (IS) metric sensitivity, with results shown for both data generating mechanisms ($X \in \mathbb{R}^1$ and $X \in \mathbb{R}^2$). It appears learning rate (LR) has the largest first order and total effects. First order effects are generally small for the other hyperparameters, but each has much larger total effects, indicating there are interactions between parameters affecting the outcome.}
\centering
\begin{tabular}[t]{lrrrrrrr}
\toprule
         & $\log_{10}(\gamma)$ & $\log_{10}(\sigma_0)$ & \makecell{Optimizer\\Steps} & NN Features & \makecell{Samples for\\Integration} & $\log_{10}$(LR) & $\mu_{0}$\\
\midrule
\addlinespace[0.3em]
\multicolumn{8}{l}{\textbf{KL, RMSE}}\\
\hspace{1em}\cellcolor{gray!10}{$X \in \mathbb{R}^1$} & \cellcolor{gray!10}{0.09(0.39)} & \cellcolor{gray!10}{0.01(0.36)} & \cellcolor{gray!10}{0.08(0.42)} & \cellcolor{gray!10}{0.04(0.36)} & \cellcolor{gray!10}{0.03(0.30)} & \cellcolor{gray!10}{0.23(0.69)} & \cellcolor{gray!10}{0.01(0.26)}\\
\hspace{1em}$X \in \mathbb{R}^2$ & 0.01(0.46) & 0.05(0.61) & 0.04(0.56) & 0.03(0.51) & 0.02(0.46) & 0.13(0.81) & 0.00(0.42)\\
\addlinespace[0.3em]
\multicolumn{8}{l}{\textbf{KL, IS}}\\
\hspace{1em}\cellcolor{gray!10}{$X \in \mathbb{R}^1$} & \cellcolor{gray!10}{0.04(0.29)} & \cellcolor{gray!10}{0.03(0.47)} & \cellcolor{gray!10}{0.06(0.45)} & \cellcolor{gray!10}{0.04(0.35)} & \cellcolor{gray!10}{0.02(0.27)} & \cellcolor{gray!10}{0.19(0.74)} & \cellcolor{gray!10}{0.01(0.25)}\\
\hspace{1em}$X \in \mathbb{R}^2$ & 0.01(0.18) & 0.04(0.40) & 0.05(0.41) & 0.04(0.44) & 0.01(0.23) & 0.12(0.76) & 0.01(0.37)\\
\bottomrule
\end{tabular}
\end{table}

\begin{table}
\centering
\caption{\label{tab:arenyitable}First order and total sensitivity indices, displayed as first order (total), for each tuning parameter when $\alpha -$Renyi divergence is used for variational inference; same format as Table~\ref{tab:kltable} which gave results for KL divergence. Similar to the KL divergence results, learning rate (LR) appears to have the highest first order and total sensitivity indices, with other hyperparameters having small first order but in some cases fairly large total sensitivity, indicating interactions between variables affecting the outcome.}
\centering
\begin{tabular}[t]{lrrrrrrr}
\toprule
         & $\alpha$ & $\log_{10}(\sigma_0)$ & \makecell{Optimizer\\Steps} & NN Features & \makecell{Samples for\\Integration} & $\log_{10}$(LR) & $\mu_{0}$\\
\midrule
\addlinespace[0.3em]
\multicolumn{8}{l}{\textbf{$\alpha -$Renyi, RMSE}}\\
\hspace{1em}\cellcolor{gray!10}{$X \in \mathbb{R}^1$} & \cellcolor{gray!10}{0.00(0.30)} & \cellcolor{gray!10}{0.00(0.30)} & \cellcolor{gray!10}{0.01(0.50)} & \cellcolor{gray!10}{0.01(0.53)} & \cellcolor{gray!10}{0.03(0.51)} & \cellcolor{gray!10}{0.12(0.94)} & \cellcolor{gray!10}{0.01(0.32)}\\
\hspace{1em}$X \in \mathbb{R}^2$ & 0.01(0.58) & 0.01(0.57) & 0.01(0.71) & 0.03(0.78) & 0.02(0.69) & 0.09(0.88) & 0.01(0.62)\\
\addlinespace[0.3em]
\multicolumn{8}{l}{\textbf{$\alpha -$Renyi, IS}}\\
\hspace{1em}\cellcolor{gray!10}{$X \in \mathbb{R}^1$} & \cellcolor{gray!10}{0.00(0.65)} & \cellcolor{gray!10}{0.01(0.67)} & \cellcolor{gray!10}{0.00(0.65)} & \cellcolor{gray!10}{0.02(0.69)} & \cellcolor{gray!10}{0.03(0.79)} & \cellcolor{gray!10}{0.13(0.93)} & \cellcolor{gray!10}{0.01(0.69)}\\
\hspace{1em}$X \in \mathbb{R}^2$ & 0.01(0.67) & 0.01(0.47) & 0.00(0.47) & 0.01(0.54) & 0.03(0.76) & 0.12(0.89) & 0.01(0.60)\\
\bottomrule
\end{tabular}
\end{table}

However, the bulk of the variability is explained in the total sensitivity indices, as most of the hyperparameters tend to have small main effects.
This indicates that in isolation they do not influence the outcome significantly, but they do influence the outcome \textit{in combination} with the other hyperparameters through interaction effects. 
For a single hyperparameter, we note that total sensitivity is often significantly different for RMSE and interval score. 
For example, for KL divergence (Table~\ref{tab:kltable}), the total sensitivity of RMSE to \(\log_{10}(\sigma_0)\) for \(X\in \mathbb{R}^1\) is 0.36 and \(X\in \mathbb{R}^2\) is 0.61, while the total sensitivity of interval score is 0.47 for \(X \in \mathbb{R}^1\) and 0.4 for \(X \in \mathbb{R}^2\). 
However, there do not appear to be consistent effects when comparing across data generating mechanisms or performance metrics.
Finally, there are some trends in magnitude across the tuning parameters. 
Total sensitivity indices for \(\mu_0\) when using KL divergence to estimate the variational fit are lower then their \(\alpha \)-Renyi counterparts, as well as lower than the total sensitivity of \(\log_{10}(\sigma_0)\) for both variational inference methods. Optimizer steps and interactions explains more of the variance when using \(\alpha \)-Renyi divergence than when using KL divergence.  The same relationship can be seen for neural network weights and samples obtained to evaluate any Monte Carlo integral within the loss function.

\begin{figure}
\centering
\includegraphics[width=0.9\linewidth]{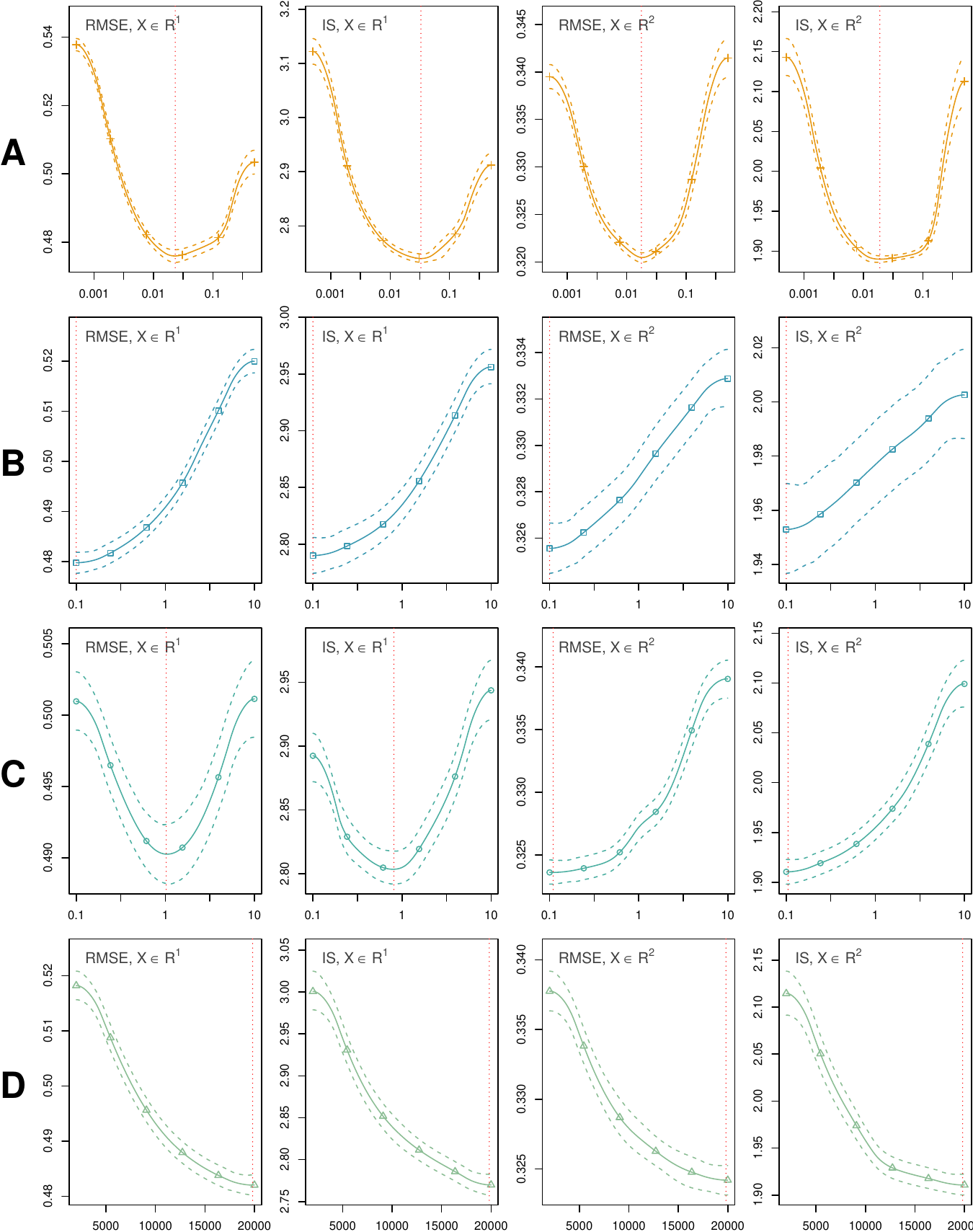}
\caption{Main effects on RMSE and IS when KL divergence is used for variational inference; each row contains results for each combination of metric (RMSE and IS) and data generating mechanism ($X\in\mathbb{R}^1$ or $X\in\mathbb{R}^2$). Each row corresponds to a different hyperparameter setting: A) \(\log_{10}(\mathrm{Learning \ Rate})\), B) KL multiplier \(\gamma\), C) prior variance \(\sigma_0^2\), and D) optimizer steps. The expected main effect is shown as the solid line, and main effect uncertainty related to the surrogate model is shown as dashed lines of the same color. The minimum of the main effect is shown as a vertical red dashed line. A) indicates that there is an ideal learning rate and that both RMSE and IS increase for smaller or larger learning rates. B) indicates that small KL multipliers result in the lowest RMSE and IS. C) shows that for $X \in \mathbb{R}^1$, prior variance near 1 is ideal, while for $X \in \mathbb{R}^2$, a smaller prior variance is preferred. D) shows that generally, large numbers of optimizer steps yield better results.}
\label{fig:maineff_kl}
\end{figure}

\begin{figure}
\centering
\includegraphics[width=0.9\linewidth]{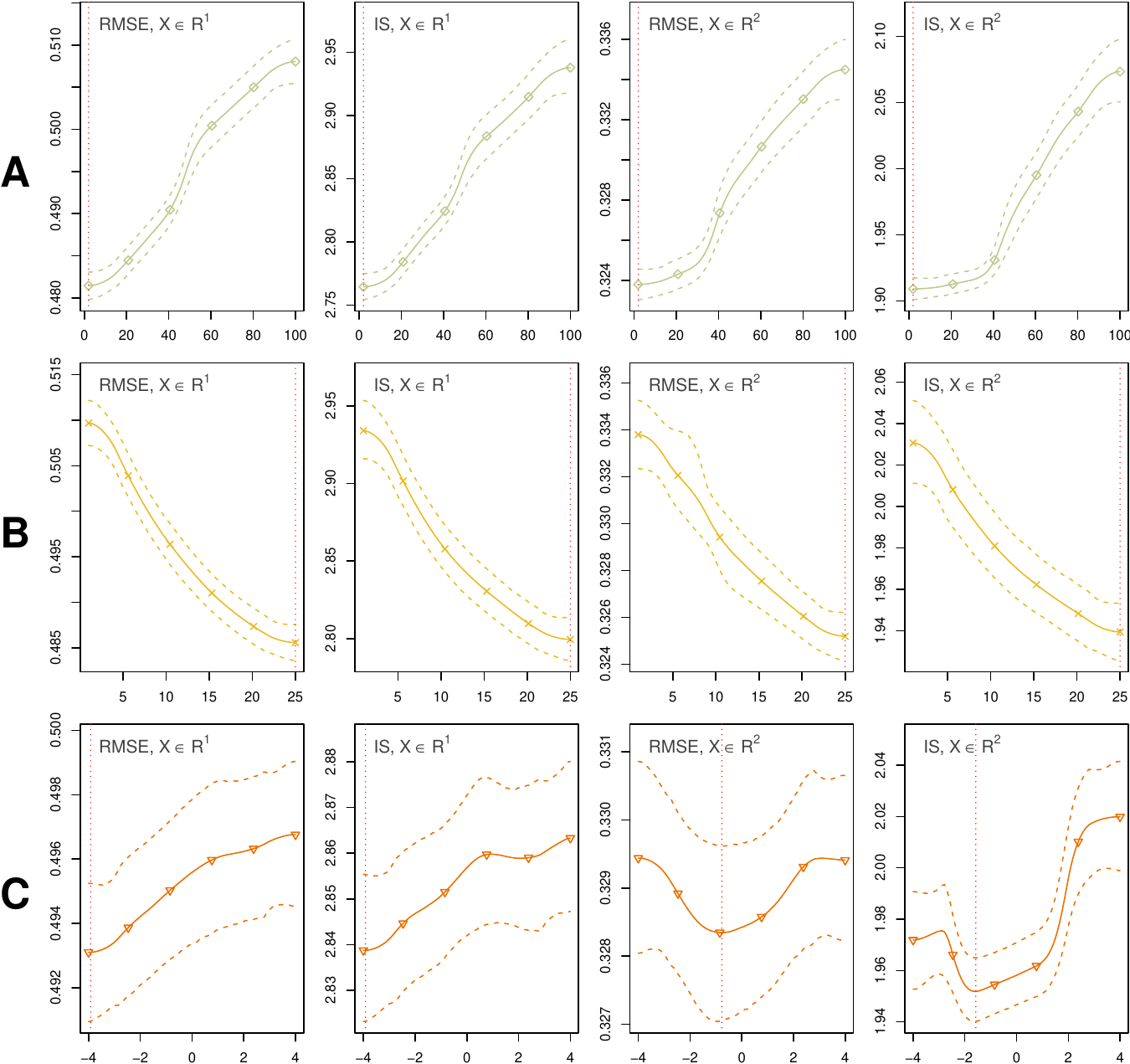}
\caption{Similar to Fig.~\ref{fig:maineff_kl} but for hyperparameters A) neural network features, B) samples for integration, and C) prior mean \(\mu_0\) when KL divergence is used for variational inference. The expected main effect is shown as the solid line, and main effect uncertainty related to the surrogate model is shown as dashed lines of the same color with the minimum of the main effect is shown as a vertical red dashed line. A) indicates that smaller numbers of neural network features were preferred in these analyses, while B) indicates that more samples for integration were beneficial. C) shows a differential effect across data generating mechanisms, with different prior means preferred in each case, though the uncertainty is relatively high.}
\label{fig:maineff_kl2}
\end{figure}

\begin{figure}
\centering
\includegraphics[width=0.9\linewidth]{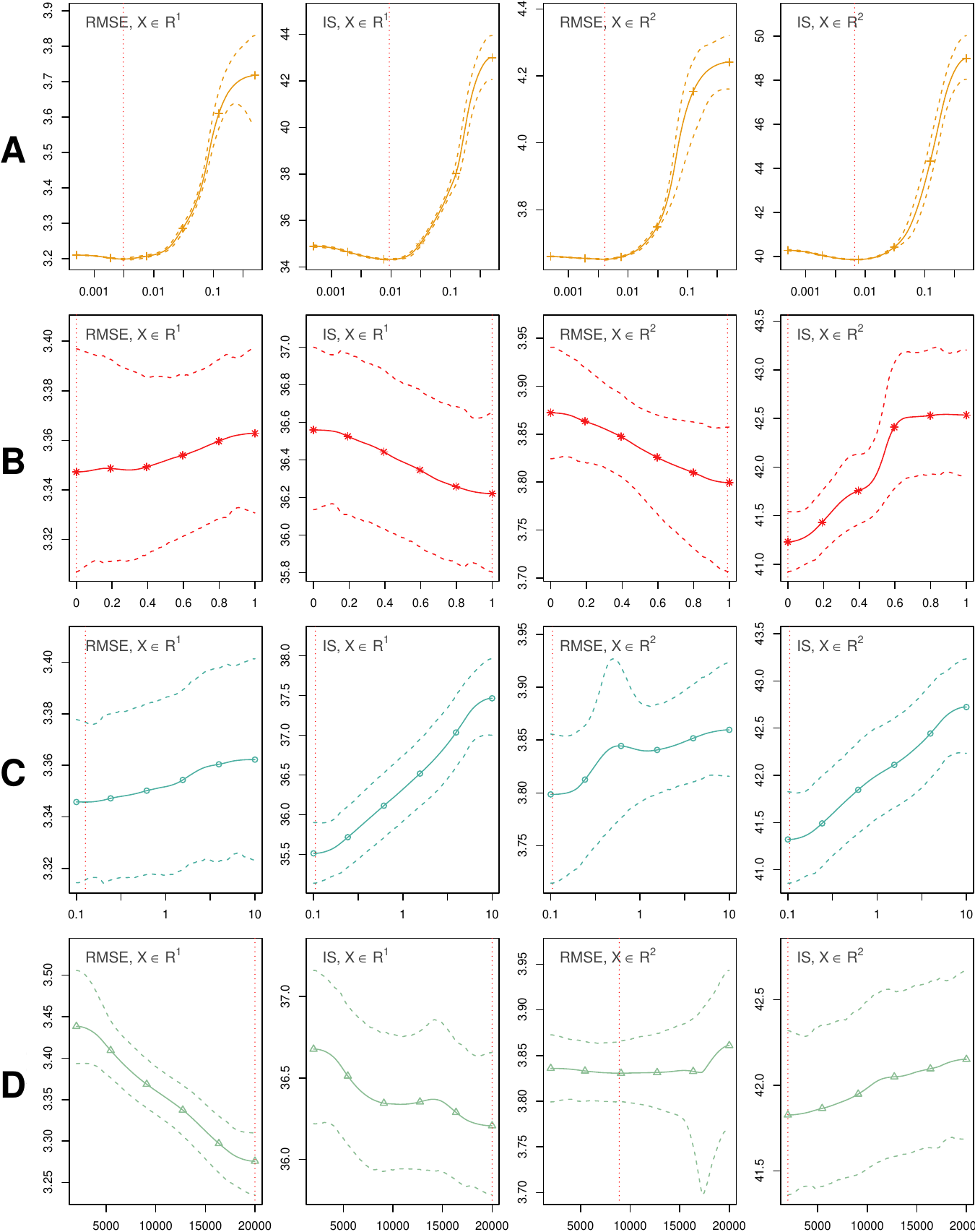}
\caption{Similar to Fig.~\ref{fig:maineff_kl}, but for $\alpha$-Renyi divergence. The rows correspond to hyperparameters A) \(\log_{10}(\mathrm{Learning \ Rate})\), B) \(\alpha\) parameter, C) prior variance \(\sigma_0^2\), and D) optimizer steps. The expected main effect is shown as the solid line, and main effect uncertainty related to the surrogate model is shown as dashed lines of the same color with the minimum of the main effect is shown as a vertical red dashed line. A) indicates that too large a learning rate leads to worse performance, but the metrics are less sensitive to a range of smaller learning rates. B) shows that the $\alpha$ parameter generally has weak effects with high uncertainty, though it does appear that for $X \in \mathbb{R}^2$, $\alpha$ near 1 is preferred for RMSE while $\alpha$ near 0 is preferred for IS. C) shows that while in some casts, uncertainty is high, generally smaller prior variances are preferable. D) shows that for $X \in \mathbb{R}^1$, more optimizer steps are better, but it seems performance decreases with more optimizer steps for $X \in \mathbb{R}^2$.}
\label{fig:maineff_arenyi}
\end{figure}

\begin{figure}
\centering
\includegraphics[width=0.9\linewidth]{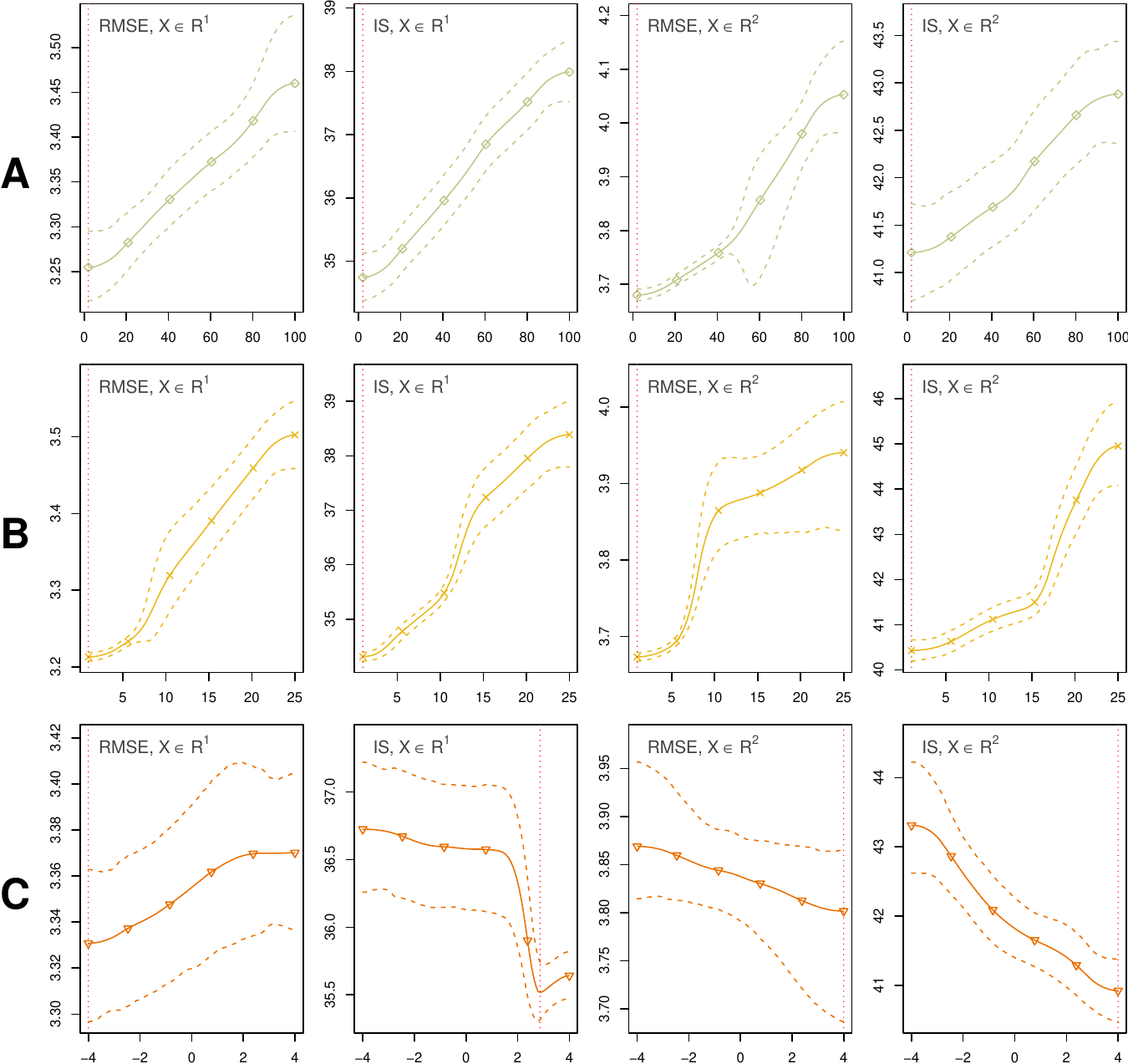}
\caption{Similar to Fig.~\ref{fig:maineff_kl}, but for $\alpha$-Renyi divergence. The rows correspond to hyperparameters A) neural network features, B) samples for integration, and C) prior mean \(\mu_0\). The expected main effect is shown as the solid line, and main effect uncertainty related to the surrogate model is shown as dashed lines of the same color with the minimum of the main effect is shown as a vertical red dashed line. A) indicates that smaller numbers of neural network features are preferred while B) indicates that smaller numbers of samples for integration are preferred. Similar to KL, the results for prior mean shown in C) are less clear, but with some evidence that results differ for the two data generating mechanisms despite higher uncertainty.}
\label{fig:maineff_arenyi2}
\end{figure}

In the remainder of this section, we discuss each hyperparameter in turn; detailed analysis of the main effects of each hyperparameter is based on Figs.~\ref{fig:maineff_kl} and ~\ref{fig:maineff_kl2} (for KL divergence) and Figs.~\ref{fig:maineff_arenyi} and ~\ref{fig:maineff_arenyi2} (for $\alpha$-Renyi divergence).
In these figures, solid lines represent the expected main effect, and dashed lines represent the resulting 5\% and 95\% quantiles for the main effect, with the hyperparameter value achieving the best metric value shown with a vertical red dashed line.
The first two columns of the matrix of plots are results for the \(X \in \mathbb{R}^1\) data generating mechanism and the second two columns correspond to the \(X \in \mathbb{R}^2\) data generating mechanism.  Columns 1 and 3 plot the main effect on RMSE, while columns 2 and 4 plot the main effect on interval score.

\paragraph{Learning rate}
As discussed above, learning rate appears to have the largest main and total effects on performance across data generating mechanisms and choice of loss function.
Fig.~\ref{fig:maineff_kl}.A shows that for KL divergence, there is a preferred learning rate which varies slightly by data generating mechanism; using learning rates that are too small or too large results in clearly worse performance according to both metrics.
For $\alpha$-Renyi divergence (Fig.~\ref{fig:maineff_arenyi}.A), there does appear to be a preferred learning rate with larger learning rates leading to worse performance, but there is perhaps more tolerance for smaller learning rates (in that they can still result in very similar performance).

\paragraph{KL multiplier} 
The main effect of the KL multiplier $\gamma$ (used only with the KL divergence loss function) is shown in Fig.~\ref{fig:maineff_kl}.B; universally across data generating mechanisms and metrics, a smaller KL multiplier results in better performance.
This is interesting because there is little theoretical backing for using $\gamma < 1$, which has the effect of downweighting the KL divergence between the prior and the variational posterior in Eq.~\ref{eq:klobjective} relative to the expected likelihood term.

\paragraph{$\alpha$-Renyi $\alpha$ value} 
Fig.~\ref{fig:maineff_arenyi}.B shows the main effect of $\alpha$ in the $\alpha$-Renyi divergence.
There is generally high uncertainty, and in the $X \in \mathbb{R}^1$ data generating mechanism, it is not clear that any particular value results in better performance.
Interestingly though in $X \in \mathbb{R}^2$, there does appear to be a differential effect in which $\alpha$ near 1 (which is most similar to KL divergence) results in the lowest RMSE while $\alpha$ near 0 results in the lowest interval score. 
These results indicate that different choices of $\alpha$ could in a sense prioritize different aspects of the resulting fit (matching the mean function value well versus characterizing the uncertainty). 

\paragraph{Prior variance} 
For KL divergence (Fig.~\ref{fig:maineff_kl}.C), there is a differential effect of prior variance $\sigma_0$ across data generating mechanisms, with $\sigma_0$ near 1 ideal for $X \in \mathbb{R}^1$ and $\sigma_0$ near the minimum value preferred for $X \in \mathbb{R}^2$, indicating that the choice of $\sigma_0$ that yields the best performance for KL divergence inference may depend heavily on the particular data set.
For $\alpha$-Renyi divergence (Fig.~\ref{fig:maineff_arenyi}.C), it appears that smaller values are preferred across the board, though there is higher uncertainty on the main effects, particularly for RMSE. 
These results could indicate that perhaps $\alpha$-Renyi is somewhat less sensitive to the choice of $\sigma_0$ compared to KL divergence, though further study is required to draw this conclusion in general across different data sets.

\paragraph{Optimizer steps} 
Increasing the number of optimizer steps is clearly beneficial for KL divergence (Fig.~\ref{fig:maineff_kl}.D). However, the effect is less clear for $\alpha$-Renyi divergence (Fig.~\ref{fig:maineff_arenyi}.D); for $X \in \mathbb{R}^1$, a larger number of steps does appear beneficial, but for $X \in \mathbb{R}^2$, this is not the case (though there is also higher uncertainty). 
This could potentially indicate overfitting with too many optimizer steps, as it appears that RMSE and IS both increase somewhat with more iterations, suggesting that as is usual practice for neural network fitting, assessing convergence by monitoring training and validation loss curves is essential.

\paragraph{Neural network features} 
In our experiments, smaller numbers of neural network features were preferred across both metrics, both data generating mechanisms, and both loss functions (KL divergence, Fig.~\ref{fig:maineff_kl2}.A; $\alpha$-Renyi, Fig.~\ref{fig:maineff_arenyi2}.A). 
This could be due to the relatively simple data generating mechanisms we used, suggesting that large numbers of features are not necessary and could lead to overfitting. 

\paragraph{Samples for integration} 
Both loss functions depend on sampling to compute expected values that are not analytically tractable; generally speaking, increasing the number of samples used to compute these values should lower the variance of the estimate and thereby introduce less noise into the stochastic gradient descent algorithm.
Generally, a larger number of samples is not used due to diminishing returns for increased computational cost, but our results do show that KL divergence does benefit from a larger number of samples (Fig.~\ref{fig:maineff_kl2}.B).
However, this is decidedly not the case for $\alpha$-Renyi divergence (Fig.~\ref{fig:maineff_arenyi2}.B), suggesting that perhaps this loss function benefits from extra stochasticity during training to locate good local optima.

\paragraph{Prior mean} 
Typically, neural network weights are initialized using distributions symmetric around 0, and it is not common to see BNN implementations use anything but a zero-mean prior on the weights.
However, our study does find that nonzero prior mean could be beneficial; for KL divergence (Fig.~\ref{fig:maineff_kl2}.C), nonzero values do appear to work well in general, though there is a fair amount of uncertainty in the effect.
Similar results are observed for $\alpha$-Renyi (Fig.~\ref{fig:maineff_arenyi2}.C).
It is worth noting that our data is standardized to be positive (in the range $[0,1]$) and we use tanh activation functions, suggesting that negative weights may be particularly useful for attaining outputs in the full range of the activation function. 
This could potentially explain why in KL divergence, negative prior means are generally preferred, though this pattern does not hold for $\alpha$-Renyi.
In general, it is difficult to interpret the sign of the weights and we expect that results could differ significantly for different data generating mechanisms or neural network architectures, but it is worth nothing that perhaps the default prior mean of zero is not always ideal.

\section{Discussion}\label{sec:discussion}


Our sensitivity analysis results quantify what we have experienced, what peers have expressed, and what can be found in the literature: successful VI for BNNs is not as straightforward as gradient-based inference for standard NNs, and common practice is more like an art than a science~\cite{tuningplaybookgithub}. 
In standard non-Bayesian NN training, the primary hyperparameters are neural network architecture (here parameterized by number of features), number of optimizer steps, and learning rate, and our results suggest that all of these are still very important for BNNs.
However, there are many additional choices that must be made for BNNs (choice of statistical divergence, divergence-specific hyperparameters such as $\gamma$ or $\alpha$, prior hyperparameters, and number of samples for Monte Carlo integration). 
Furthermore, BNNs must be assessed by not only their predictive accuracy (e.g., RMSE), but also by the quality of their uncertainty intervals (which we measure here using IS).

Our results show that most of the variability in RMSE and IS is explained through \textit{interactions} between the various hyperparameters, making it difficult to give simple prescriptions for BNN success.
While our current results cannot elucidate the exact nature of the second- and higher-order interactions, we have some intuition about possible pairwise interactions to consider in future work.
For example, in KL divergence, we expect an interaction between $\gamma$ and the prior variance $\sigma_0^2$, as the KL divergence between the prior and the variational posterior will be affected by $\sigma_0$ and this term in the loss function is then weighted by $\gamma$.
We also expect there to be an interaction between learning rate and number of optimizer steps needed for convergence, as in standard NN training.
In addition, the prior variance could interact with the complexity of the network architecture; for a given Gaussian prior with variance $\sigma_0^2$ on each weight and an input comprised of all ones, the output of the layer (pre-activation) will have variance $n \sigma_0^2$ where $n$ is the number of weights, so variance downscaling may be beneficial as the number of weights increases.
We plan to further probe these and other possible interactions in future work.

In terms of first-order effects, KL divergence consistently had the largest sum of first order indices for both data generating mechanisms and performance metrics when compared to using \(\alpha -\)Renyi divergence.  
Results will vary depending on the data generating mechanism, training data, and the uncertainty distributions used in the sensitivity analysis, but these results indicate that a \textit{BNN artisan} can effect change on the quality of their BNN fit (for better or worse) when adjusting one parameter at a time more readily when using KL divergence than when using \(\alpha -\)Renyi divergence.
When interpreting any values from the main effects, we have to consider the fact that the hyperparameters corresponding to minima are the computed when averaging over all other hyperparameters, so the minima identified across the set of hyperparameters may not actually correspond to the best performing model.  
For example, there will be specific tuning parameter settings where the minimum main effect learning rate does not produce the minimum RMSE or IS.
However, these values, their order of magnitude, and their consistency (or lack there of) between data generating mechanisms and performance metrics can provide a starting point for fine tuning.

When comparing the tuning parameters individually, \(\log_{10}(\mathrm{LR})\) stands out as having the largest influence on all results with the highest first order and total sensitivity indices.  This is a quantification of what is anecdotally known:  ``\textit{Learning rate is important.}''.  The first order sensitivity index for \(\log_{10}(\mathrm{LR})\) makes up roughly half of the sum of the first order sensitivity indices over all parameters for a given data generating mechanism and performance metric.  
The main effect plots (Figs.~\ref{fig:maineff_kl} and \ref{fig:maineff_arenyi}) show relatively low uncertainty of the main effect, which could be attributed to the more clear and independent mapping of learning rate to an observed RMSE and IS. 
The minimum main effect of \(\log_{10}(\mathrm{LR})\) appears to be similar when comparing all data generating mechanisms and performance metrics, given a statistical distance. 
When comparing the main effect function between RMSE and IS, the trend appears to be similar, with slight differences around the minimum value and between the boundaries of an individual plot.  
These results suggest that even if other hyperparameters are set using heuristics, tuning the learning rate is essential for attaining good BNN performance.

When using KL divergence for variational inference, our main effect results show the average RMSE or IS is minimized for the smallest choice of \(\gamma\) we considered, equal to 0.1.  When evaluating our loss function in Eq.~\ref{eq:klobjective}, \(\gamma\) downweights the influence of the prior density for the neural network weights.  
Downweighting the prior allows the expectation of the likelihood with respect to the variational posterior to have a larger influence, and allows the variational posterior to move further away from an almost certainly misspecified prior.  
Further research could investigate if there is similar consistency for optimal values of \(\gamma\) for more complicated data generating mechanisms, as well as if even smaller values of \(\gamma\) are on average optimal.
We were surprised to see little clear trend for the main effect of choosing \(\alpha\) when using \(\alpha \)-Renyi divergence for estimating the variational posterior.  The literature points to a relation between specific values of \(\alpha\) and differing statistical distances, including KL divergence for \(\alpha \to 1\) and Hellinger distance for \(\alpha = 0.5\)~\cite{li2016renyi}. 
The expected main effects have little variation, reflected in the first order indices, and relatively large uncertainties.  
We do not claim that the choice of \(\alpha\) is irrelevant, as these results are affected by the hyperrectangle of parameter choices considered for the analysis, but there is no clear choice of \(\alpha\) from these results. 
Interestingly, in the case of the $X \in \mathbb{R}^2$ data generating mechanism, the two metrics (RMSE and IS) appeared to prefer very different $\alpha$ values, suggesting that some choices of $\alpha$ may be better for predictive accuracy while others may be better for uncertainty quantification.
As has been noted in previous works, $\alpha$-Renyi divergence may be more robust to prior misspecification when $\alpha \neq 1$~\cite{knoblauch2022optimization}. 
Indeed, prior hyperparameters are difficult to select using \textit{a priori} knowledge~\cite{fortuinbayesian}, and our results show that performance across divergences and data generating mechanisms is affected by prior variance (and to a lesser degree prior mean), so other hyperparameters that minimize the effect of a misspecified prior may be important to tune. 


Given the results presented here, we make the following suggestions for users of BNNs.
\begin{itemize}
    \item Systematic tuning of the hyperparameters for BNNs is essential, regardless of the form of the divergence used in the loss function. While grid search is simple to implement, more advanced methods such as randomized search or Bayesian optimization~\cite{schonlau1997computer,klein2017fast} can reduce the computational cost of tuning. 
    \item With a very large computational budget, producing main effect plots as part of fitting a BNN can help diagnose which tuning parameters matter, and which ranges of tuning parameters to consider. Sensitivity analysis has previously been used to reduce the dimensionality of an optimization problem~\cite{taddy2009bayesian}. However, incorporating a sensitivity analysis into every BNN fit is computationally expensive, especially for more complex and higher dimensional data.
    \item Users should keep in mind that multiple metrics (including RMSE and IS, among others) may be helpful for tuning as they may not always agree on the best hyperparameter settings.
    \item Standard hyperparameters for neural networks, including learning rate, number of optimizer steps, and architecture are important to tune for BNNs, with learning rate showing the largest first order and total effects, so these should be prioritized if there is limited time to explore all hyperparameters, but the user should be aware that BNN performance is also sensitive to complex interactions between other settings (including prior hyperparameters).
    \item Prior hyperparameters are difficult to specify in a principled manner, so instead focusing on tuning other hyperparameters that may minimize the impact of a misspecified prior (such as $\gamma$ or $\alpha$) may be a suitable approach.
\end{itemize}

In summary, our global sensitivity analysis reveals that careful and targeted hyperparameter tuning is crucial for optimizing BNN performance. 
While some hyperparameters like the learning rate consistently exhibit strong influence on RMSE and interval score, others interact in subtle ways that can significantly affect predictive accuracy and UQ. 
Therefore, leveraging sensitivity analysis or related surrogate-based methods such as Bayesian optimization to guide the tuning process can offer valuable insights into model behavior. 
Ultimately, balancing principled tuning with practical constraints may enable more robust and valuable usage BNNs in applications where predictive uncertainty is key.


\section{Acknowledgements}

This manuscript has been authored with number LA-UR-25-26090 by Triad National Security under Contract with the U.S. Department of Energy.  Research presented in this article was supported by the Laboratory Directed Research and Development program of Los Alamos National Laboratory under project number 20230469ECR.

Code utilized in this research was developed with the help of chatGPT GPT-4o to improve productivity and aid in debugging.

\section{Declaration of Interest Statement}

There are no competing interests to declare.

\section{References}
\vspace{-0.75cm}
\renewcommand{\refname}{}
\bibliographystyle{unsrt}
\bibliography{main}

\appendix
\setcounter{table}{0} \renewcommand{\thetable}{A\arabic{table}}
\setcounter{figure}{0} \renewcommand{\thefigure}{A\arabic{figure}}

\section{Parameters of Best and Worst BNN Fits}

\begin{table}[!h]
\centering
\caption{\label{tab:bestworst}Parameter settings for the best and worst observed fits from the \(X \in \mathbb{R}^1\) data generating mechanism.  All numeric values are displayed on the natural scale to make reading easier.  }
\centering
\resizebox{\ifdim\width>\linewidth\linewidth\else\width\fi}{!}{
\begin{tabular}[t]{llrrrrrrrrrr}
\toprule
Method & Category & RMSE & IS & $\gamma$ & $\alpha$ & $\sigma_0$ & \makecell{Optimizer\\Steps} & \makecell{NN\\Features} & \makecell{Samples for\\Integration} & LR & $\mu_{0}$\\
\midrule
\cellcolor{gray!10}{KL} & \cellcolor{gray!10}{Best IS} & \cellcolor{gray!10}{0.20} & \cellcolor{gray!10}{0.81} & \cellcolor{gray!10}{0.10} & \cellcolor{gray!10}{NA} & \cellcolor{gray!10}{0.71} & \cellcolor{gray!10}{10821} & \cellcolor{gray!10}{6} & \cellcolor{gray!10}{5} & \cellcolor{gray!10}{0.0226} & \cellcolor{gray!10}{-0.97}\\
KL & Worst IS & 1.39 & 30.31 & 4.36 & NA & 1.70 & 2611 & 97 & 6 & 0.0006 & -1.25\\
\cellcolor{gray!10}{KL} & \cellcolor{gray!10}{Best RMSE} & \cellcolor{gray!10}{0.19} & \cellcolor{gray!10}{0.83} & \cellcolor{gray!10}{0.16} & \cellcolor{gray!10}{NA} & \cellcolor{gray!10}{1.67} & \cellcolor{gray!10}{17030} & \cellcolor{gray!10}{61} & \cellcolor{gray!10}{6} & \cellcolor{gray!10}{0.0075} & \cellcolor{gray!10}{-0.08}\\
KL & Worst RMSE & 2.56 & 6.52 & 6.65 & NA & 1.27 & 13869 & 99 & 13 & 0.2489 & 0.57\\
\addlinespace
\cellcolor{gray!10}{$\alpha -$Renyi} & \cellcolor{gray!10}{Best IS} & \cellcolor{gray!10}{0.22} & \cellcolor{gray!10}{0.88} & \cellcolor{gray!10}{NA} & \cellcolor{gray!10}{0.87} & \cellcolor{gray!10}{1.22} & \cellcolor{gray!10}{16744} & \cellcolor{gray!10}{4} & \cellcolor{gray!10}{16} & \cellcolor{gray!10}{0.0075} & \cellcolor{gray!10}{-0.13}\\
$\alpha -$Renyi & Worst IS & 26.72 & 1570.32 & NA & 0.30 & 1.10 & 3778 & 65 & 21 & 0.4787 & -1.39\\
\cellcolor{gray!10}{$\alpha -$Renyi} & \cellcolor{gray!10}{Best RMSE} & \cellcolor{gray!10}{0.20} & \cellcolor{gray!10}{1.12} & \cellcolor{gray!10}{NA} & \cellcolor{gray!10}{0.39} & \cellcolor{gray!10}{2.33} & \cellcolor{gray!10}{16455} & \cellcolor{gray!10}{29} & \cellcolor{gray!10}{2} & \cellcolor{gray!10}{0.1559} & \cellcolor{gray!10}{-0.36}\\
$\alpha -$Renyi & Worst RMSE & 136.61 & 612.38 & NA & 0.31 & 0.77 & 2748 & 42 & 15 & 0.4925 & 1.10\\
\bottomrule
\end{tabular}}
\end{table}

\section{Tables with captions on individual pages}

\renewcommand{\thetable}{\arabic{table}}
\renewcommand{\thefigure}{\arabic{figure}}

\setcounter{table}{0}

\begin{table}[!h]
\centering
\caption{\label{tab:expparams}Bounds for resulting goodness of fit data collection as well as integration for global sensitivity analysis.  For a given statistical distance (KL or $\alpha$-Renyi), the bounds shown are used with both the $X \in \mathbb{R}^1$ and $X \in \mathbb{R}^2$ data generating mechanisms.}
\centering
\begin{tabular}[t]{lrrrrrrrr}
\toprule
    & $\log_{10}(\gamma)$ & $\alpha$ & $\log_{10}(\sigma_0)$ & \makecell{Optimizer\\Steps} & NN Features & \makecell{Samples for\\Integration} & $\log_{10}$(LR) & $\mu_{0}$\\
\midrule
\addlinespace[0.3em]
\multicolumn{9}{l}{\textbf{KL Parameters}}\\
\hspace{1em}\cellcolor{gray!10}{Upper Bound} & \cellcolor{gray!10}{1} & \cellcolor{gray!10}{NA} & \cellcolor{gray!10}{0.5} & \cellcolor{gray!10}{20000} & \cellcolor{gray!10}{100} & \cellcolor{gray!10}{25} & \cellcolor{gray!10}{-0.3} & \cellcolor{gray!10}{2}\\
\hspace{1em}Lower Bound & -1 & NA & -0.5 & 2000 & 2 & 1 & -3.3 & -2\\
\addlinespace[0.3em]
\multicolumn{9}{l}{\textbf{$\alpha -$Renyi Parameters}}\\
\hspace{1em}\cellcolor{gray!10}{Upper Bound} & \cellcolor{gray!10}{NA} & \cellcolor{gray!10}{1} & \cellcolor{gray!10}{0.5} & \cellcolor{gray!10}{20000} & \cellcolor{gray!10}{100} & \cellcolor{gray!10}{25} & \cellcolor{gray!10}{-0.3} & \cellcolor{gray!10}{2}\\
\hspace{1em}Lower Bound & NA & 0 & -0.5 & 2000 & 2 & 1 & -3.3 & -2\\
\bottomrule
\end{tabular}
\end{table}

\newpage

\begin{table}[!h]
\centering
\caption{\label{tab:kltable} First order and total sensitivity indices, displayed as first order (total), for each tuning parameter when KL divergence is used for variational inference; top section corresponds to the RMSE metric sensitivity while bottom section corresponds to the interval score (IS) metric sensitivity, with results shown for both data generating mechanisms ($X \in \mathbb{R}^1$ and $X \in \mathbb{R}^2$). It appears learning rate (LR) has the largest first order and total effects. First order effects are generally small for the other hyperparameters, but each has much larger total effects, indicating there are interactions between parameters affecting the outcome.}
\centering
\begin{tabular}[t]{lrrrrrrr}
\toprule
         & $\log_{10}(\gamma)$ & $\log_{10}(\sigma_0)$ & \makecell{Optimizer\\Steps} & NN Features & \makecell{Samples for\\Integration} & $\log_{10}$(LR) & $\mu_{0}$\\
\midrule
\addlinespace[0.3em]
\multicolumn{8}{l}{\textbf{KL, RMSE}}\\
\hspace{1em}\cellcolor{gray!10}{$X \in \mathbb{R}^1$} & \cellcolor{gray!10}{0.09(0.39)} & \cellcolor{gray!10}{0.01(0.36)} & \cellcolor{gray!10}{0.08(0.42)} & \cellcolor{gray!10}{0.04(0.36)} & \cellcolor{gray!10}{0.03(0.30)} & \cellcolor{gray!10}{0.23(0.69)} & \cellcolor{gray!10}{0.01(0.26)}\\
\hspace{1em}$X \in \mathbb{R}^2$ & 0.01(0.46) & 0.05(0.61) & 0.04(0.56) & 0.03(0.51) & 0.02(0.46) & 0.13(0.81) & 0.00(0.42)\\
\addlinespace[0.3em]
\multicolumn{8}{l}{\textbf{KL, IS}}\\
\hspace{1em}\cellcolor{gray!10}{$X \in \mathbb{R}^1$} & \cellcolor{gray!10}{0.04(0.29)} & \cellcolor{gray!10}{0.03(0.47)} & \cellcolor{gray!10}{0.06(0.45)} & \cellcolor{gray!10}{0.04(0.35)} & \cellcolor{gray!10}{0.02(0.27)} & \cellcolor{gray!10}{0.19(0.74)} & \cellcolor{gray!10}{0.01(0.25)}\\
\hspace{1em}$X \in \mathbb{R}^2$ & 0.01(0.18) & 0.04(0.40) & 0.05(0.41) & 0.04(0.44) & 0.01(0.23) & 0.12(0.76) & 0.01(0.37)\\
\bottomrule
\end{tabular}
\end{table}

\newpage

\begin{table}[!h]
\centering
\caption{\label{tab:arenyitable}First order and total sensitivity indices, displayed as first order (total), for each tuning parameter when $\alpha -$Renyi divergence is used for variational inference; same format as Table~\ref{tab:kltable} which gave results for KL divergence. Similar to the KL divergence results, learning rate (LR) appears to have the highest first order and total sensitivity indices, with other hyperparameters having small first order but in some cases fairly large total sensitivity, indicating interactions between variables affecting the outcome.}
\centering
\begin{tabular}[t]{lrrrrrrr}
\toprule
         & $\alpha$ & $\log_{10}(\sigma_0)$ & \makecell{Optimizer\\Steps} & NN Features & \makecell{Samples for\\Integration} & $\log_{10}$(LR) & $\mu_{0}$\\
\midrule
\addlinespace[0.3em]
\multicolumn{8}{l}{\textbf{$\alpha -$Renyi, RMSE}}\\
\hspace{1em}\cellcolor{gray!10}{$X \in \mathbb{R}^1$} & \cellcolor{gray!10}{0.00(0.30)} & \cellcolor{gray!10}{0.00(0.30)} & \cellcolor{gray!10}{0.01(0.50)} & \cellcolor{gray!10}{0.01(0.53)} & \cellcolor{gray!10}{0.03(0.51)} & \cellcolor{gray!10}{0.12(0.94)} & \cellcolor{gray!10}{0.01(0.32)}\\
\hspace{1em}$X \in \mathbb{R}^2$ & 0.01(0.58) & 0.01(0.57) & 0.01(0.71) & 0.03(0.78) & 0.02(0.69) & 0.09(0.88) & 0.01(0.62)\\
\addlinespace[0.3em]
\multicolumn{8}{l}{\textbf{$\alpha -$Renyi, IS}}\\
\hspace{1em}\cellcolor{gray!10}{$X \in \mathbb{R}^1$} & \cellcolor{gray!10}{0.00(0.65)} & \cellcolor{gray!10}{0.01(0.67)} & \cellcolor{gray!10}{0.00(0.65)} & \cellcolor{gray!10}{0.02(0.69)} & \cellcolor{gray!10}{0.03(0.79)} & \cellcolor{gray!10}{0.13(0.93)} & \cellcolor{gray!10}{0.01(0.69)}\\
\hspace{1em}$X \in \mathbb{R}^2$ & 0.01(0.67) & 0.01(0.47) & 0.00(0.47) & 0.01(0.54) & 0.03(0.76) & 0.12(0.89) & 0.01(0.60)\\
\bottomrule
\end{tabular}
\end{table}

\newpage

\section{Figures}

\begin{figure}[!h]
\centering
\includegraphics[width=\linewidth]{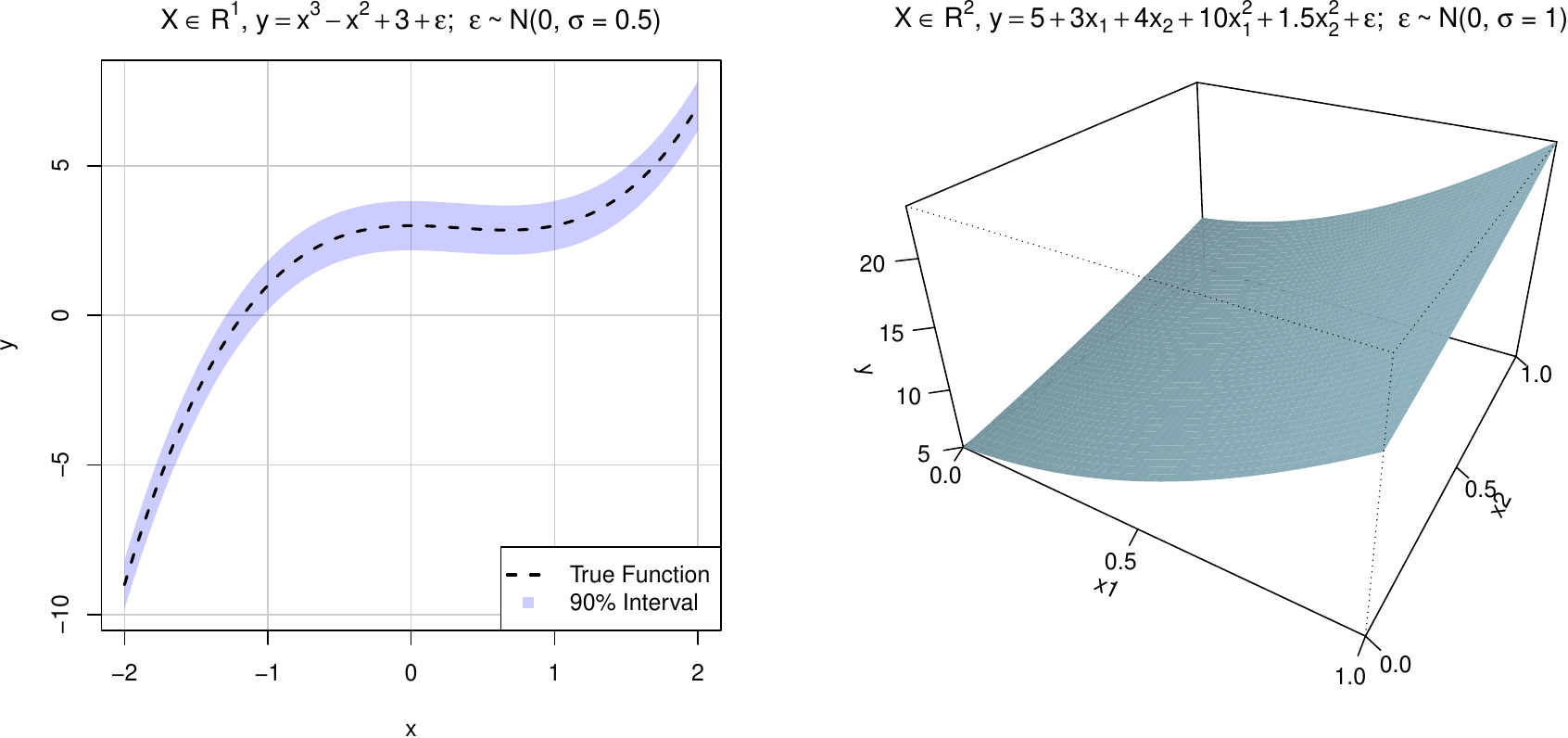}
\caption{The data generating mechanisms used for testing the variational BNN approximations.  For the \(X \in \mathbb{R}^1\) data generating mechanism (left) the expected value of the function is plotted along with the 90\% conditional distribution interval arising from additive homoskedastic noise.  For the \(X \in \mathbb{R}^2\) data generating mechanism (right), we show the expected value of the function with respect to the two input dimensions.}
\end{figure}

\newpage

\begin{figure}[!h]
\centering
\begin{adjustbox}{width=1.05\linewidth, center}
\includegraphics[trim=0.1cm 0.05cm 0.05cm 0.1cm, clip]{figs/fig2.pdf}
\end{adjustbox}
\caption{BNN variational approximations through minimizing KL divergence.  The best and worst resulting fits from the 750 different initializations used to collect data are shown. The best fits were selected as either the minimum observed RMSE or IS, and the worst were selected as either the largest observed RMSE or IS. The best fits according to interval score and RMSE are qualitatively similar and appear to describe the data distribution accurately. The worst fit according to interval score appears flat with very high variance, while the worst fit according to RMSE incurs high bias and moderately high variance.} 
\end{figure}

\newpage

\begin{figure}[!h]
\centering
\includegraphics[width=0.9\linewidth]{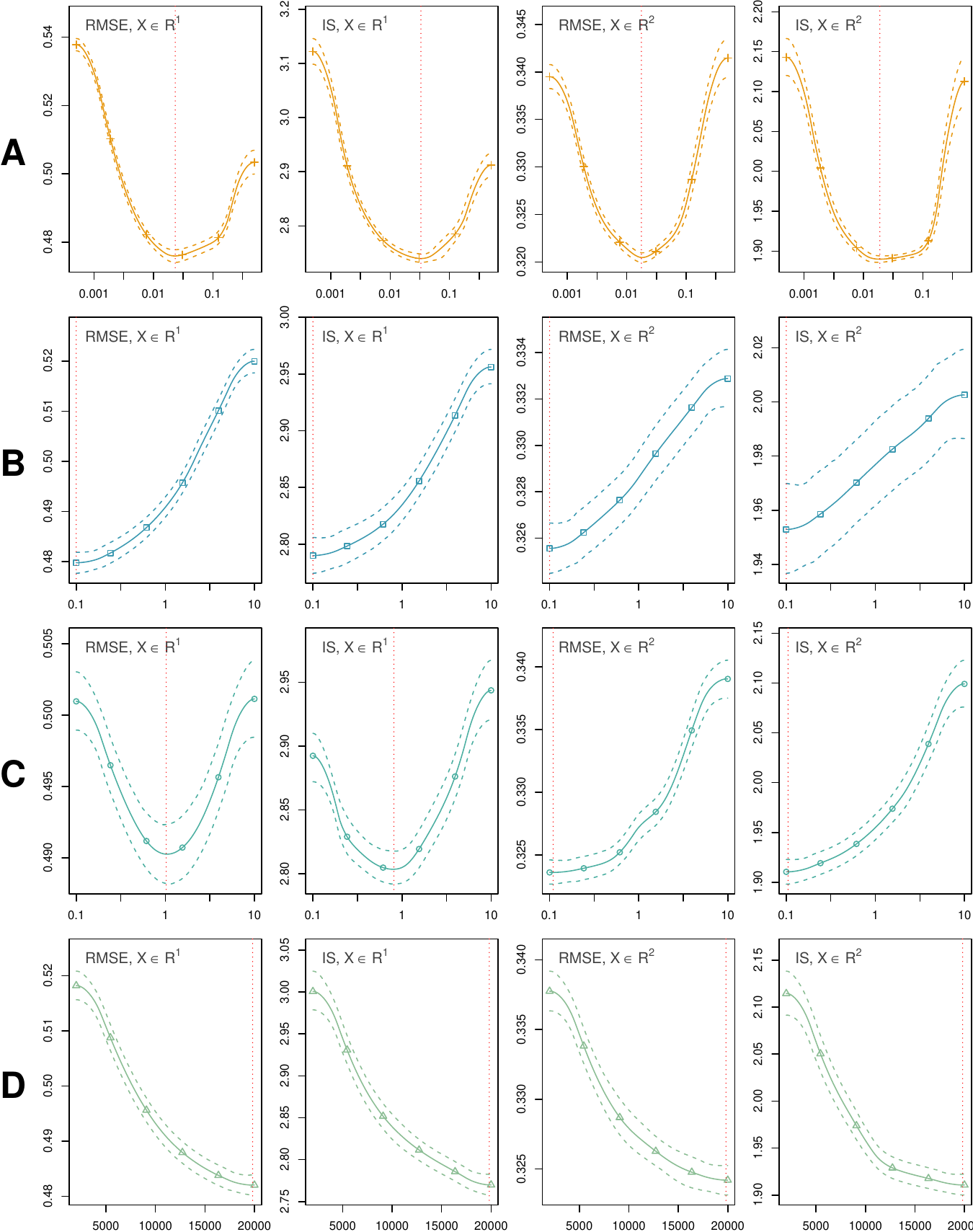}
\caption{Main effects on RMSE and IS when KL divergence is used for variational inference; each row contains results for each combination of metric (RMSE and IS) and data generating mechanism ($X\in\mathbb{R}^1$ or $X\in\mathbb{R}^2$). Each row corresponds to a different hyperparameter setting: A) \(\log_{10}(\mathrm{Learning \ Rate})\), B) KL multiplier \(\gamma\), C) prior variance \(\sigma_0^2\), and D) optimizer steps. The expected main effect is shown as the solid line, and main effect uncertainty related to the surrogate model is shown as dashed lines of the same color. The minimum of the main effect is shown as a vertical red dashed line. A) indicates that there is an ideal learning rate and that both RMSE and IS increase for smaller or larger learning rates. B) indicates that small KL multipliers result in the lowest RMSE and IS. C) shows that for $X \in \mathbb{R}^1$, prior variance near 1 is ideal, while for $X \in \mathbb{R}^2$, a smaller prior variance is preferred. D) shows that generally, large numbers of optimizer steps yield better results.}
\end{figure}

\newpage

\begin{figure}[!h]
\centering
\includegraphics[width=\linewidth]{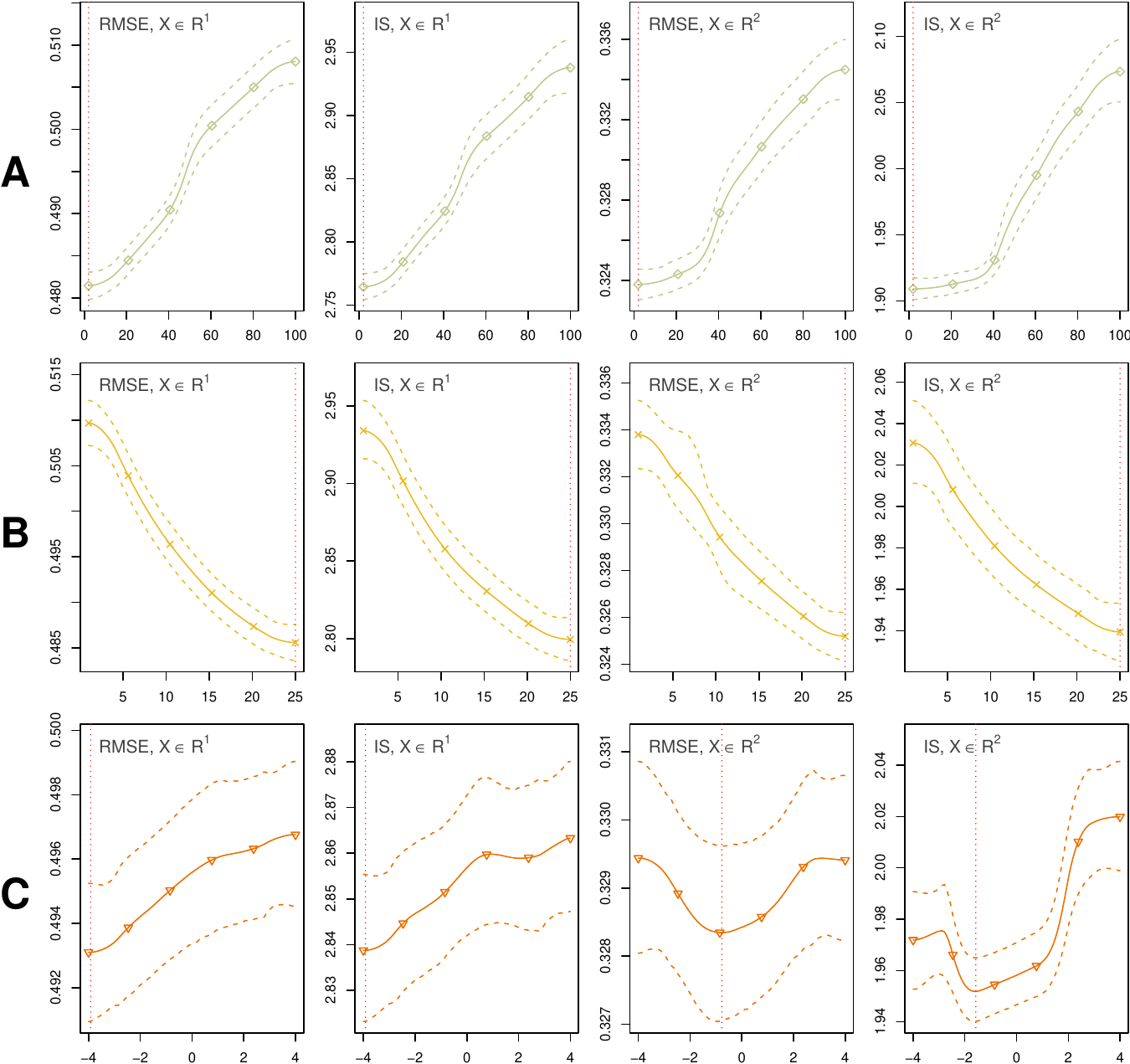}
\caption{Similar to Fig.~\ref{fig:maineff_kl} but for hyperparameters A) neural network features, B) samples for integration, and C) prior mean \(\mu_0\) when KL divergence is used for variational inference. The expected main effect is shown as the solid line, and main effect uncertainty related to the surrogate model is shown as dashed lines of the same color with the minimum of the main effect is shown as a vertical red dashed line. A) indicates that smaller numbers of neural network features were preferred in these analyses, while B) indicates that more samples for integration were beneficial. C) shows a differential effect across data generating mechanisms, with different prior means preferred in each case, though the uncertainty is relatively high.}
\end{figure}

\newpage

\begin{figure}[!h]
\centering
\includegraphics[width=0.9\linewidth]{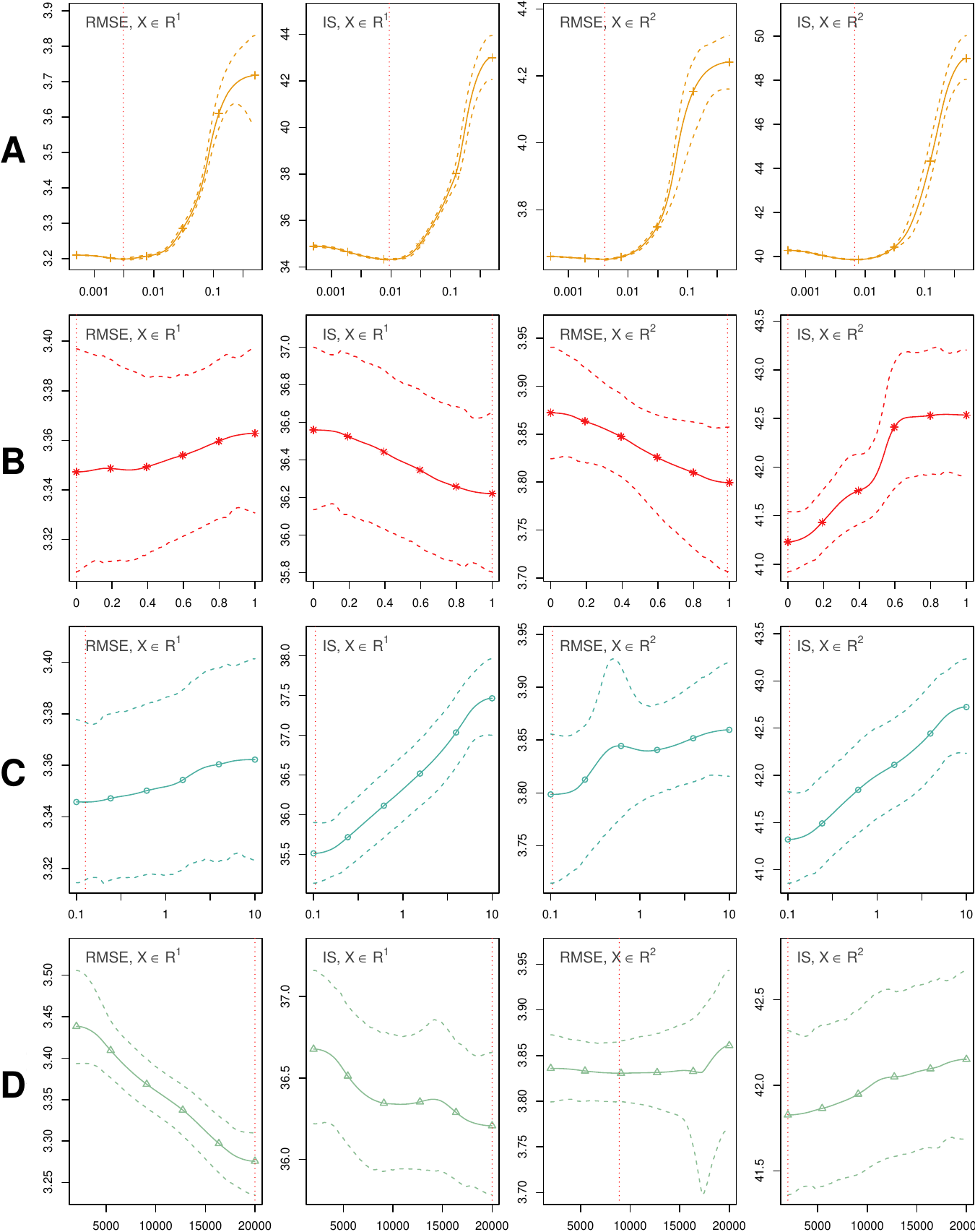}
\caption{Similar to Fig.~\ref{fig:maineff_kl}, but for $\alpha$-Renyi divergence. The rows correspond to hyperparameters A) \(\log_{10}(\mathrm{Learning \ Rate})\), B) \(\alpha\) parameter, C) prior variance \(\sigma_0^2\), and D) optimizer steps. The expected main effect is shown as the solid line, and main effect uncertainty related to the surrogate model is shown as dashed lines of the same color with the minimum of the main effect is shown as a vertical red dashed line. A) indicates that too large a learning rate leads to worse performance, but the metrics are less sensitive to a range of smaller learning rates. B) shows that the $\alpha$ parameter generally has weak effects with high uncertainty, though it does appear that for $X \in \mathbb{R}^2$, $\alpha$ near 1 is preferred for RMSE while $\alpha$ near 0 is preferred for IS. C) shows that while in some casts, uncertainty is high, generally smaller prior variances are preferable. D) shows that for $X \in \mathbb{R}^1$, more optimizer steps are better, but it seems performance decreases with more optimizer steps for $X \in \mathbb{R}^2$.}
\end{figure}

\newpage

\begin{figure}[!h]
\centering
\includegraphics[width=\linewidth]{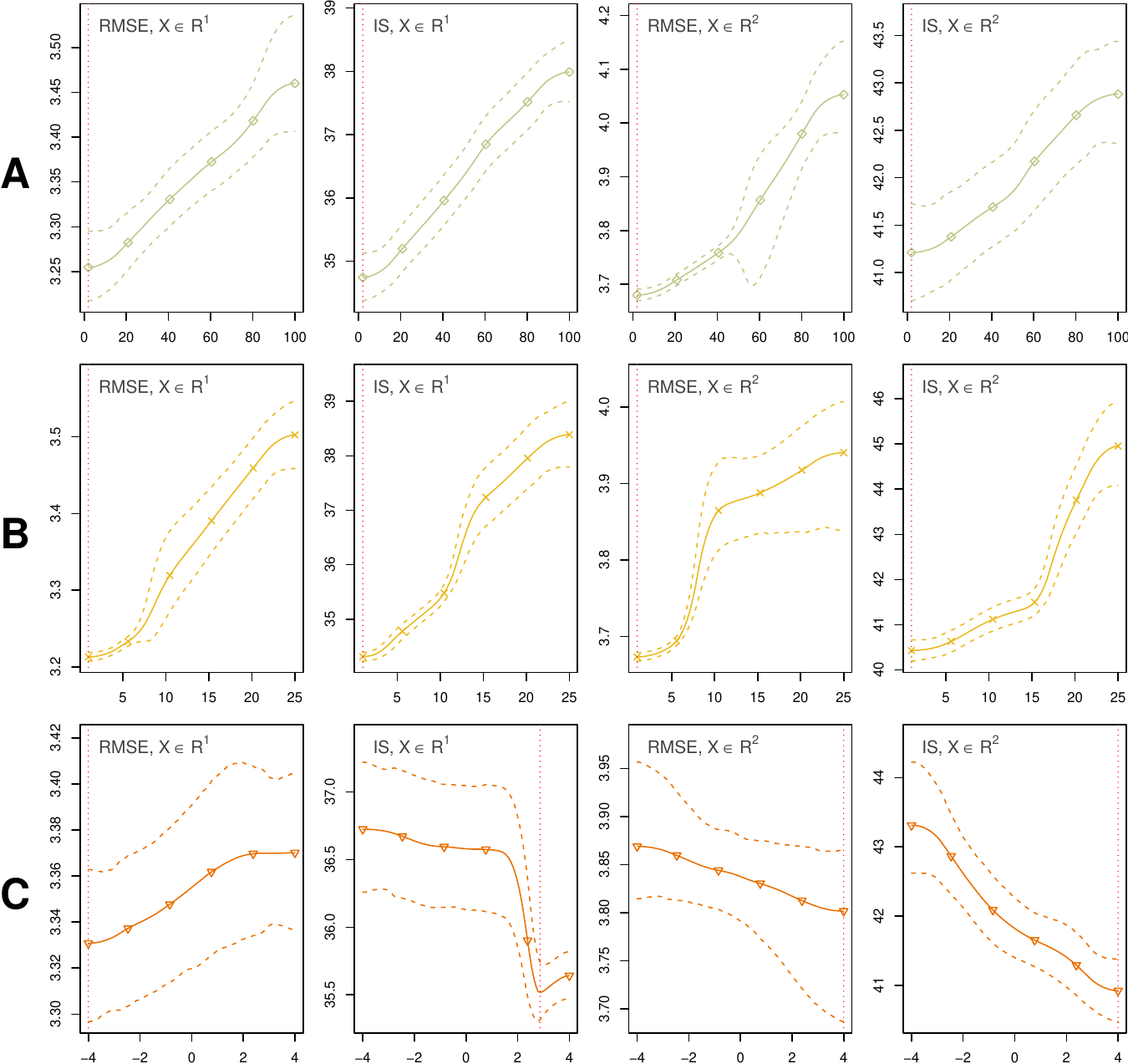}
\caption{Similar to Fig.~\ref{fig:maineff_kl}, but for $\alpha$-Renyi divergence. The rows correspond to hyperparameters A) neural network features, B) samples for integration, and C) prior mean \(\mu_0\). The expected main effect is shown as the solid line, and main effect uncertainty related to the surrogate model is shown as dashed lines of the same color with the minimum of the main effect is shown as a vertical red dashed line. A) indicates that smaller numbers of neural network features are preferred while B) indicates that smaller numbers of samples for integration are preferred. Similar to KL, the results for prior mean shown in C) are less clear, but with some evidence that results differ for the two data generating mechanisms despite higher uncertainty.}
\end{figure}

\clearpage

\section{Figure Captions as a list}

List of figure captions:

\begin{enumerate}
\item The data generating mechanisms used for testing the variational BNN approximations.  For the \(X \in \mathbb{R}^1\) data generating mechanism (left) the expected value of the function is plotted along with the 90\% conditional distribution interval arising from additive homoskedastic noise.  For the \(X \in \mathbb{R}^2\) data generating mechanism (right), we show the expected value of the function with respect to the two input dimensions.\\

\item BNN variational approximations through minimizing KL divergence.  The best and worst resulting fits from the 750 different initializations used to collect data are shown. The best fits were selected as either the minimum observed RMSE or IS, and the worst were selected as either the largest observed RMSE or IS. The best fits according to interval score and RMSE are qualitatively similar and appear to describe the data distribution accurately. The worst fit according to interval score appears flat with very high variance, while the worst fit according to RMSE incurs high bias and moderately high variance. \\

\item Main effects on RMSE and IS when KL divergence is used for variational inference; each row contains results for each combination of metric (RMSE and IS) and data generating mechanism ($X\in\mathbb{R}^1$ or $X\in\mathbb{R}^2$). Each row corresponds to a different hyperparameter setting: A) \(\log_{10}(\mathrm{Learning \ Rate})\), B) KL multiplier \(\gamma\), C) prior variance \(\sigma_0^2\), and D) optimizer steps. The expected main effect is shown as the solid line, and main effect uncertainty related to the surrogate model is shown as dashed lines of the same color. The minimum of the main effect is shown as a vertical red dashed line. A) indicates that there is an ideal learning rate and that both RMSE and IS increase for smaller or larger learning rates. B) indicates that small KL multipliers result in the lowest RMSE and IS. C) shows that for $X \in \mathbb{R}^1$, prior variance near 1 is ideal, while for $X \in \mathbb{R}^2$, a smaller prior variance is preferred. D) shows that generally, large numbers of optimizer steps yield better results. \\

\item Similar to Fig.~\ref{fig:maineff_kl} but for hyperparameters A) neural network features, B) samples for integration, and C) prior mean \(\mu_0\) when KL divergence is used for variational inference. The expected main effect is shown as the solid line, and main effect uncertainty related to the surrogate model is shown as dashed lines of the same color with the minimum of the main effect is shown as a vertical red dashed line. A) indicates that smaller numbers of neural network features were preferred in these analyses, while B) indicates that more samples for integration were beneficial. C) shows a differential effect across data generating mechanisms, with different prior means preferred in each case, though the uncertainty is relatively high. \\

\item Similar to Fig.~\ref{fig:maineff_kl}, but for $\alpha$-Renyi divergence. The rows correspond to hyperparameters A) \(\log_{10}(\mathrm{Learning \ Rate})\), B) \(\alpha\) parameter, C) prior variance \(\sigma_0^2\), and D) optimizer steps. The expected main effect is shown as the solid line, and main effect uncertainty related to the surrogate model is shown as dashed lines of the same color with the minimum of the main effect is shown as a vertical red dashed line. A) indicates that too large a learning rate leads to worse performance, but the metrics are less sensitive to a range of smaller learning rates. B) shows that the $\alpha$ parameter generally has weak effects with high uncertainty, though it does appear that for $X \in \mathbb{R}^2$, $\alpha$ near 1 is preferred for RMSE while $\alpha$ near 0 is preferred for IS. C) shows that while in some casts, uncertainty is high, generally smaller prior variances are preferable. D) shows that for $X \in \mathbb{R}^1$, more optimizer steps are better, but it seems performance decreases with more optimizer steps for $X \in \mathbb{R}^2$. \\

\item Similar to Fig.~\ref{fig:maineff_kl}, but for $\alpha$-Renyi divergence. The rows correspond to hyperparameters A) neural network features, B) samples for integration, and C) prior mean \(\mu_0\). The expected main effect is shown as the solid line, and main effect uncertainty related to the surrogate model is shown as dashed lines of the same color with the minimum of the main effect is shown as a vertical red dashed line. A) indicates that smaller numbers of neural network features are preferred while B) indicates that smaller numbers of samples for integration are preferred. Similar to KL, the results for prior mean shown in C) are less clear, but with some evidence that results differ for the two data generating mechanisms despite higher uncertainty.

\end{enumerate}

\end{document}